%% file: main.tex
\documentclass[11pt,a4paper]{article}
\usepackage{times,latexsym}
\usepackage{url}
\usepackage[T1]{fontenc}

\usepackage[acceptedWithA]{tacl2021v1}
\usepackage{xspace,mfirstuc,tabulary}

\usepackage{amsmath}
\usepackage{nccmath}
\usepackage{amssymb}
\usepackage{amsfonts}
\usepackage{float}
\usepackage{bbm} % for \mathbbm command
\usepackage{xcolor}
\usepackage{graphicx} % For \resizebox
\usepackage[inline]{enumitem}
\usepackage{booktabs, multirow} % for borders and merged ranges

\definecolor{olivegreen}{RGB}{107,142,35}

\definecolor{psychedelicpurple}{RGB}{142,122,181}

\usepackage{algpseudocode}

\usepackage{booktabs,arydshln}
\makeatletter
\def\adl@drawiv#1#2#3{%
        \hskip.5\tabcolsep
        \xleaders#3{#2.5\@tempdimb #1{1}#2.5\@tempdimb}%
                #2\z@ plus1fil minus1fil\relax
        \hskip.5\tabcolsep}
\newcommand{\cdashlinelr}[1]{%
  \noalign{\vskip\aboverulesep
           \global\let\@dashdrawstore\adl@draw
           \global\let\adl@draw\adl@drawiv}
  \cdashline{#1}
  \noalign{\global\let\adl@draw\@dashdrawstore
           \vskip\belowrulesep}}
\makeatother

\usepackage{listings}
\usepackage{pifont}
\newcommand{\cmark}{\ding{51}}

\usepackage{bm}

\usepackage[linesnumbered,ruled,vlined]{algorithm2e}
\definecolor{commentcolor}{RGB}{110,154,155} % define comment color
\newif\iftaclinstructions
\taclinstructionsfalse % AUTHORS: do NOT set this to true
\iftaclinstructions
\renewcommand{\confidential}{}
\renewcommand{\anonsubtext}{(No author info supplied here, for consistency with
TACL-submission anonymization requirements)}
\newcommand{\instr}
\fi

\iftaclpubformat % this "if" is set by the choice of options

\else

\fi

\title{Adversarial Defence without \textit{Adversarial Defence}: Enhancing Language Model Robustness via Instance-level Principal Component Removal}

\author{
Yang Wang$^{1,4}$\quad Chenghao Xiao$^2$\quad Yizhi Li$^1$\quad \\ \textbf{Stuart E. Middleton}$^3$\quad \textbf{Noura Al Moubayed}$^2$\quad \textbf{Chenghua Lin}$^1$\thanks{\, \, Corresponding author.}
   \\
   $^1$The University of Manchester, UK\quad $^2$Durham University, UK \\
   $^3$The University of Southampton, UK\quad $^4$Automated Analytics, UK \\
    \texttt{yang.wang-27@postgrad.manchester.ac.uk} \\
    \texttt{chenghua.lin@manchester.ac.uk}
}

\date{}

\begin{document}
\maketitle

\begin{abstract}
Pre-trained language models (PLMs) have driven substantial progress in natural language processing but remain vulnerable to adversarial attacks, raising concerns about their robustness in real-world applications. 
Previous studies have sought to mitigate the impact of adversarial attacks by introducing adversarial perturbations into the training process, either implicitly or explicitly. 
While both strategies enhance robustness, they often incur high computational costs. 
In this work, we propose a simple yet effective add-on module that enhances the adversarial robustness of PLMs by removing instance-level principal components, without relying on conventional adversarial defences or perturbing the original training data. 
Our approach transforms the embedding space to approximate Gaussian properties, thereby reducing its susceptibility to adversarial perturbations while preserving semantic relationships. 
This transformation aligns embedding distributions in a way that minimises the impact of adversarial noise on decision boundaries, enhancing robustness without requiring adversarial examples or costly training-time augmentation. 
Evaluations on eight benchmark datasets show that our approach improves adversarial robustness while maintaining comparable before-attack accuracy to baselines, achieving a balanced trade-off between robustness and generalisation. 
\end{abstract}

% \begin{center}
% \vspace{0.03cm}
% \begin{tabular}{c@{\hskip 0.2cm}l}
%     \raisebox{-.25\height}{\href{https://github.com/PuReDefence/PuRe}{\includegraphics[width=0.4cm]{figures/github-mark}}} & {\small\texttt{\href{https://github.com/PuReDefence/PuRe}{github.com/PuReDefence/PuRe}}}
% \end{tabular}
% \vspace{0.03cm}
% \end{center}

% \vspace{-0.6cm} % adjust until spacing looks right
\begin{center}
\begin{tabular}{c@{\hskip 0.2cm}l}
    \raisebox{-.25\height}{\href{https://github.com/PuReDefence/PuRe}{\includegraphics[width=0.4cm]{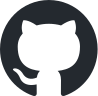}}} & {\small\texttt{\href{https://github.com/PuReDefence/PuRe}{github.com/PuReDefence/PuRe}}}
\end{tabular}
\end{center}

\section{Introduction}
\label{sec:introduction}

Pre-trained language models (PLMs) have exhibited remarkable performance across various fields such as computer vision~\cite{dosovitskiy2021an, pmlr-v139-touvron21a, khan2022transformers, wang2023internimage, zhu2024vision} and natural language processing (NLP)~\cite{devlin2018bert, liu2019roberta, yang2019xlnet, he2021deberta, he2021debertav3, asl2023robustembed}. Although they have achieved great success in a number of fields, their vulnerability to adversarial attacks has unveiled a significant challenge to models' robustness by adding small human-imperceptible perturbations to normal examples \cite{sun2020adv, li2020bert, he2021model, jha2023codeattack}.

Existing adversarial defence methods often demand extensive computational resources, or have limited improvements in adversarial robustness. 
For example, adversarial training-based methods \cite{madry2018towards, Zhu2020FreeLB:, li2021token, wang2021infobert} involve generating perturbations through multiple iterations during training, which significantly increases the computational overhead. 
Similarly, some ensemble-based techniques leverage the statistical properties of the ensemble to provably certify the robustness \cite{ye-etal-2020-safer, zhou-etal-2021-defense, moon-etal-2023-randomized, zeng2023certified}, leading to additional costs during both training and inference. 
An alternative line of defence leverages regularisation-based methods \cite{10.5555/3524938.3525366, wang2021infobert, liu2022flooding, yang-etal-2023-domain}, which are more computationally efficient but tend to show limited improvements in robustness against adversarial attacks \cite{zhu-rao-2023-exploring}. 
This disparity highlights the need for more efficient defence methods that strike a balance between computational efficiency and robustness enhancement.

To address these challenges, we propose Purified Representation (\textsc{PuRe}) to enhance adversarial robustness without introducing adversarial perturbations during training, either implicitly or explicitly\footnote{The title is intended to be paradoxical. It signifies that \textsc{PuRe} enhances adversarial robustness without employing the common strategies of adversarial defence (e.g., generating adversarial examples for training).}. 
% Specifically, \textsc{PuRe} is applied via standard fine-tuning of the PLM with the module integrated and access to the model's architecture and gradients. 
\textsc{PuRe} is implemented as a module that is integrated directly into the PLM's architecture. 
The entire model is then trained using a standard fine-tuning process, requiring no special modifications. 
At its core, this module leverages Principal Component Removal \cite{arora2017simple} to reshape the embedding space. 
By removing dominant components, it encourages representations to align more closely with Gaussian-like distributions, which reduces the model's sensitivity to the targeted perturbations that adversaries often exploit. 
This transformation strengthens robustness without relying on adversarial example generation or resource-intensive training augmentations, providing an efficient and practical solution for improving adversarial resilience in NLP tasks. 
% The entire model is then trained using a standard fine-tuning process, requiring no special modifications. 
% It then leverages Principal Component Removal \cite{arora2017simple}, a technique that reshapes the embedding space by removing dominant components, thereby encouraging representations to align more closely with Gaussian-like distributions while preserving semantic relationships. 
% This transformation strengthens robustness against adversarial perturbations without relying on adversarial example generation or resource-intensive training augmentations, providing an efficient and practical solution for improving adversarial resilience in NLP tasks. 
The evaluation of \textsc{PuRe} is underpinned by benchmarking eight language understanding datasets, spanning across sentiment analysis, subjectivity status classification, paraphrase identification, textual entailment, and commonsense reasoning. 
\textsc{PuRe} shows superior textual adversarial defence ability to most tasks, while performing on-par with the baselines in terms of before-attack accuracy, indicating a good trade-off between robustness and generalisation. 
% \red{It is important to note that \textsc{PuRe} requires standard fine-tuning of the PLM with the module integrated, necessitating access to the model's architecture and gradients. }
Our contributions can be summarised as follows:
\begin{itemize}
    % \item We propose \textsc{PuRe}, a simple yet effective add-on module designed to improve adversarial resilience in NLP tasks without generating adversarial  perturbations into the training process, either implicitly or explicitly. 
    \item We introduce \textsc{PuRe}, a novel, parameter-free module for improving adversarial robustness. Its plug-and-play design allows it to be easily integrated into PLMs and optimised with standard fine-tuning, eliminating the need for costly adversarial training.
    % \item We demonstrate that increasing isotropy (i.e. have uniform variance across all dimensions in embedding space) in PLMs effectively improve performance on various models and fine-tuning tasks in adversarial settings. 
    \item We are the first to empirically demonstrate that making the embedding space more geometrically uniform via Principal Component Removal is a highly effective defence mechanism for PLMs.
    % \item We conduct extensive experiments across multiple models and benchmark tasks, demonstrating that \textsc{PuRe} significantly boosts adversarial robustness against strong attacks while maintaining performance on clean data.
\end{itemize}

\section{Related Work}

% Adversarial robustness refers to a model's ability to resist intentional perturbations crafted to exploit its weaknesses \cite{madry2018towards, alayrac2019labels, Wang2020Improving, tsai2021formalizing, wang2021towards}. 
% This differs from general robustness, which measures resilience to unintentional noise or natural distribution shifts (e.g., typos, domain mismatches). 
% While both aim to improve reliability, adversarial robustness assumes an adversary with knowledge of the model's architecture or training data, actively searching for perturbations that maximally degrade performance. 

The concept of model robustness is twofold. 
General robustness addresses resilience to natural, unintentional variations arising from real-world noise. 
Adversarial robustness, on the other hand, addresses resilience to malicious, intentional perturbations. These are carefully crafted by an adversary to be imperceptible to humans yet cause the model to fail \cite{madry2018towards, alayrac2019labels, Wang2020Improving, tsai2021formalizing, wang2021towards}. 
In this work, we focus specifically on enhancing adversarial robustness.

\subsection{Adversarial Attacks}

% Adversarial examples were initially explored by \citet{szegedy2013intriguing} in the context of computer vision, where visually imperceptible distortions to input images could significantly maximise the network's prediction error. 
% Hence, the computational demand for generating such adversarial examples encouraged the development of gradient-based techniques, including FGSM \cite{goodfellow2014explaining}, PGD \cite{madry2017towards, madry2018towards, madry2019deep}, and BIM \cite{kurakin2018adversarial}.
The field of adversarial attacks was pioneered by \citet{szegedy2013intriguing} in computer vision, where they demonstrated that visually imperceptible distortions could cause models to misclassify images with high confidence. 
The computational cost of this initial method spurred the development of more efficient gradient-based attacks, including the Fast Gradient Sign Method \cite[FGSM]{goodfellow2014explaining} and Projected Gradient Descent  \cite[PGD]{madry2018towards}.

Transferring adversarial attack methods from computer vision to NLP introduces unique challenges due to the discrete nature of textual data as opposed to continuous pixel values. Thus, NLP-focused adversarial attack research has largely focused on crafting semantics-preserving perturbations. 
For example, back-translation \cite{iyyer2018adversarial} generates adversarial examples by translating text back and forth between different languages. 
\citet{wang2020cat} use GANs \cite{goodfellow2014generative, li-etal-2020-dgst} to create fluent adversarial texts that closely resemble natural language. Additionally, methods have been developed to identify critical words in text and replace them with synonyms or to introduce character-level perturbations such as typos in letters, numbers, or special symbols \cite{jin2020bert, maheshwary2021strong, li2018textbugger}.

These advancements in adversarial attack methods have driven a deeper understanding of NLP models' vulnerabilities, motivating the development of robust defence strategies to counteract a wide range of adversarial threats.

\subsection{Adversarial Defences}

Adversarial defences in NLP aim to enhance the robustness of models against adversarial perturbations. The primary defence strategies can be classified into four categories: adversarial training-based, perturbation control-based, certification-based, and regularisation-based methods.

\noindent\textbf{Adversarial training-based methods}~~involve augmenting the training data with adversarial examples, enabling the model to learn in an environment that simulates attacks in the training process, either implicitly or explicitly \cite{jin2020bert, morris2020textattack, si2020better, wang2021infobert, hauser2023bert}. Implicit approaches usually generate perturbations dynamically in the embedding space as a part of the training process, which improves the model's resilience to a range of adversarial scenarios \cite{wu-etal-2017-adversarial, Zhu2020FreeLB:, dong2021towards, gao-etal-2023-dsrm, latorre2023finding}. 
Explicit approaches, on the other hand, involves generating adversarial examples in the input space (text data) using adversarial attack methods \cite{jin2020bert, li-etal-2020-bert-attack, tan-etal-2020-morphin, zang-etal-2020-word}, and these pre-generated adversarial examples will be incorporated into the training pipeline. 
We refer this explicit adversarial training-based approach as Adversarially-augmented (AdvAug) training (see \S\ref{sec:aat}). 
Despite its efficacy and interpretability, the adversarial training-based methods are often computationally intensive due to the need for extensive adversarial example generation and fine-tuning.

\noindent\textbf{Perturbation control-based methods}~~aim to detect and correct adversarial inputs by incorporating mechanisms to recognise potential perturbations \cite{alshemali2019toward, yoo2022detection, shen2023textshield, ali2023detect} or by altering the perturbation toward cleaner inputs to limit the adversarial space \cite{10.5555/3304222.3304371, zhang2020deep, zhou2021defense, bao2021defending}. 
Techniques include spell-checking systems for character-level defences that correct adversarially manipulated inputs before classification \cite{alshemali2019toward} and word-level defences that substitute input words with synonyms to neutralise adversarial effects \cite{ye-etal-2020-safer, zhou2021defense, dong2021towards}. 
However, synonym-based methods often face limitations in practical scenarios, where the perturbation sets of potential attacks are usually unknown \cite{li2021searching}.

\noindent\textbf{Certification-based methods}~~provide theoretical guarantees by constructing a perturbation-resistant region around the input space \cite{wang2019improving, dong2021towards, asl2023robustembed, moon2023randomized, zeng2023certified}. Although these methods offer strong theoretical assurances, they typically involve impractical constraints in real-world applications. 
Certification-based methods can require extensive computational resources and long verification times \cite{zeng2023certified}, which may not be feasible in applications with limited computational capacity or real-time processing requirements. 

\noindent\textbf{Regularisation-based methods}~~add regularisation terms to the loss function to improve model robustness without relying on adversarial examples generation or pre-defined synonym sets. 
For example, \citet{wang2021infobert} introduced two regularisers to improve out-of-domain robustness evaluated on adversarial NLI \cite{nie-etal-2020-adversarial} and SQuAD \cite{jia-liang-2017-adversarial} datasets. 
The first regulariser is an implementation of the Information Bottleneck principle \cite{tishby2015deep} specialised for contextual text representations, and the second regulariser is to minimise the mutual information between the input and the representation. 
\citet{liu2022flooding} introduced a ``flooding'' loss \cite{10.5555/3524938.3525366}, which helps models avoid overconfidence in predictions by maintaining the loss at a specific threshold. 
Their findings suggest that the flooding method shows promise in defending against adversarial attacks. 
\citet{yang-etal-2023-domain} modified the traditional label smoothing technique \cite{guo2017calibration} to account for adversarial perturbations, thereby enhancing model resilience. 
These structure-free approaches offer computational advantages over methods that depend on explicitly generated adversarial data or pre-defined perturbation sets.

\subsection{Isotropic Latent Space}

Isotropy in the context of representation learning refers to the uniform distribution of the directions of vectors in the embedding space, implying that no particular direction is overly dominant \cite{mu2018allbutthetop}. 
The embeddings spread more evenly across all dimensions, resembling a spherical Gaussian-like distribution where all directions are statistically similar. 

\citet{mu2018allbutthetop} propose a post-processing algorithm that masks out the top principal components of the data, and show that it improves performance for Word2Vec \cite{mikolov2013efficient} and GloVe \cite{pennington-etal-2014-glove} embeddings on word similarity tasks. 
Achieving an isotropic latent space has also been explored in prior work \cite{li-etal-2020-sentence, huang-etal-2021-whiteningbert-easy, su2021whitening}, arguing that improving isotropy in the embedding space improves model performance. 
Similarly, Kernel-Whitening \cite{gao-etal-2022-kernel} employs isotropic transformations to mitigate dataset bias, demonstrating the benefits of a uniform representation space for generalisation. 
More recent approaches such as I-STAR \cite{rudman2024stable}, which is a differentiable and mini-batch-stable isotropy-based regularisation scheme, studies the relationship between fine-tuned model performance and isotropy. 
Contrary to previous works in NLP, \citet{rudman2024stable} find that further decreasing isotropy improves downstream model performance. 
While these methods enhance the quality of embeddings for downstream tasks, they often serve as a post-processing step and do not explicitly address adversarial robustness. 

On the other hand, \textsc{PuRe} builds on the idea of isotropic representations but shifts the focus towards adversarial robustness. 
We hypothesise that isotropic transformation can reduce the sensitivity to adversarial perturbations and regularise decision boundaries, providing a more robust defence mechanism. 
To sum up, we derive several keys to distinguish \textsc{PuRe} from existing adversarial defence methods. 
\begin{enumerate*}[label=(\roman*)]
\item \textsc{PuRe} obviates the need for generating adversarial examples, whether implicitly or explicitly, resulting in significant computational savings. 
\item It addresses adversarial vulnerabilities via Principal Component Removal, thereby providing a robust defence mechanism that does not rely on particular attack constraints. 
\item It is a simple, add-on module that can be seamlessly integrated with off-the-shelf PLMs, offering a model-agnostic solution. 
\end{enumerate*}

\section{Purified Representation (\textsc{PuRe})}
\label{sec:Pure}

We propose \textsc{PuRe} (Purified Representation), a method designed to improve adversarial robustness by encouraging isotropy in the representation space (i.e., making embeddings more uniformly distributed across dimensions). 
This isotropic structure reduces sensitivity to adversarial perturbations and strengthens the stability of decision boundaries. 
\textsc{PuRe} achieves this through a simple yet effective adaptation of Principal Component Analysis \cite[PCA]{abdi2010principal} to standardise the latent space. 
In this section, we detail the design and intuition behind \textsc{PuRe}.

\subsection{Instance-level Principal Components Removal}
\label{sec:pcr}

The core idea behind \textsc{PuRe} is to reduce the dominance of certain directions in the representation space by removing principal components that capture most of the variance. 
Traditional PCA typically discards the weakest directions (i.e., principal components with the least variance) to minimise information loss. 
For example, BERT-whitening \cite{su2021whitening} applies PCA to BERT embeddings by discarding less informative dimensions, thereby retaining important textual features and improving performance in semantic similarity tasks. 
In contrast, \textsc{PuRe} applies PCA in a novel manner, aiming for significant information reduction to enhance adversarial robustness. 
\textsc{PuRe} subtracts these dominant components from the final layer token-level representations. 
This results in a representation space that is closer to an isotropic distribution, where all directions carry roughly equal importance (see Figure.~\ref{fig:dsv}).

\textsc{PuRe} draws inspiration from techniques like SIF embeddings \cite{arora2017simple}, which removes the top-1 principal component from static embeddings to capture variance in rogue dimensions \cite{timkey2021all}, making the representation space more isotropic. 
However, rather than applying Principal Component Removal (PCR) as a post-processing step to the entire corpus, \textsc{PuRe} performs this operation at the instance level, removing projections onto the top-1 principal component of the subspace spanned by individual tokens within a sentence during fine-tuning. 
We combined with efficient principal component computation via Singular Value Decomposition \cite[SVD]{golub1971singular}, enables end-to-end training while achieving an isotropic latent space, which is shown ultimately improving the model's resilience to adversarial perturbations. 
Preliminaries of PCA and SVD can be found in Appendix.~\ref{sec:preliminaries}.

% Typically, PCA eliminates the weakest directions, rather than the dominant directions (i.e. removing principal components starting from the least ones), so that we sacrifice as little information as possible. 
% For example, in BERT-whitening \cite{su2021whitening}, PCA retains dominant textual features by discarding less informative BERT \cite{devlin2018bert} embedding dimensions, thus preserving crucial information for tasks like semantic similarity while reducing complexity and enhancing performance.
% For our purposes, we apply PCA in a different way, to achieve as much information reduction as possible.

% Building on findings similar to SIF embeddings \cite{arora2017simple}, they remove the top-1 principal component from static embedding models, which helps capture the variance of rogue dimensions \cite{timkey2021all} in the model and makes the space more isotropic.
% Instead of applying principal component removal (PCR) to the entire set of corpus word embeddings as a post-processing technique, we remove projections onto the top-1 principal component of the spanned subspace at an instance level, focusing on the embeddings of individual tokens within a sentence during fine-tuning. 
% This approach allows for end-to-end model training.

Suppose having final layer token-level embedding $\mathbf{X} \in \mathbb{R}^{n \times d}$, with a sequence length $n$ and embedding dimension $d$. We perform SVD on $\mathbf{X}$ and get the right singular matrix $\mathbf{V}$. The columns of $\mathbf{V}$ are the corresponding principal components (since SVD directly computes the eigenvector matrix $\mathbf{V}$), which are already sorted by descending eigenvalue. 
We null away the top-1 principal component\footnote{We investigated the impact of removing the top-k principal components, and observed a plummet in before-attack accuracy. Therefore, we set the default to removing only the top-1 principal component. Ablation study can be found in \S\ref{sec:remove_top_k_pc_qwqen}.}:
\begin{equation}
    \mathbf{X} \gets \mathbf{X} - (\mathbf{X} \mathbf{v}_1) \mathbf{v}^\top_1
    \label{eq:pcr}
\end{equation}

Eq.~\ref{eq:pcr} is equivalent to removing rank-1 matrix corresponding to largest singular value from $\mathbf{X}$:
% % \begin{equation}
\begin{align}
    \mathbf{X} - (\mathbf{X} \mathbf{v}_1) \mathbf{v}^\top_1 \nonumber &= \mathbf{X} - (\mathbf{u}_1 \mathbf{\sigma}_1) \mathbf{v}^\top_1 \\
    &= (\sum\limits_{i=1}^{k} \sigma_i \mathbf{u}_i \mathbf{v}^\top_i) - \sigma_1 \mathbf{u}_1 \mathbf{v}^\top_1
\end{align}
% \end{equation}

% \begin{equation}
%     (\sum\limits_{i=1}^{k} \sigma_i \mathbf{u}_i \mathbf{v}^\top_i) - \sigma_1 \mathbf{u}_1 \mathbf{v}^\top_1
% \end{equation}

This operation essentially removes the component of  $\mathbf{X}$ that is in the direction of the largest singular value, represented by $\sigma_1 \mathbf{u}_1 \mathbf{v}^\top_1$. The largest singular value, $\sigma_1$, and its corresponding singular vectors, $\mathbf{u}_1$ and $\mathbf{v}_1$, capture the most significant mode of variation (or the principal component) in the tokens embedding matrix $\mathbf{X}$. 
Building upon the findings of \citet{DBLP:conf/iclr/MuV18}, they observe that by post-processing the word representation by eliminating the common parts, the processed word representations is able to capture stronger linguistic regularities (i.e. the semantic similarity of words is well captured by the similarity of the corresponding vector representations). They posit that PCR makes the representations more \textit{isotropic} with stronger self-normalisation properties. 
We then hypothesise that a uniform distribution of embeddings can lead to more stable decision boundaries, because adversarial attacks often seek to exploit the model by finding inputs that cross these boundaries with minimal changes. A more isotropic space might reduce the number of ``weak spots'' or vulnerabilities that adversarial inputs can exploit.
% Therefore, we hypothesise that the removal of the principal component associated with the largest singular value in each token's embedding may diminish the impact of textual adversarial attacks. 
Therefore, if the dominant principal component corresponding to the largest singular value is thought to represent noise or an unwanted signal, its removal can help in focusing on more subtle underlying structures and consequently yield a more distilled and essence-focused representation of the text.

% The final tokens embedding matrix is obtained by subtracting the projection of $\mathbf{\tilde{X}}$ to their top $r$ principal components.

% This principal components removal operation on is on dense low-dimensional tokens matrix.

\subsubsection{Randomised SVD}
\label{sec:rsvd}

Traditional methods for SVD can be computationally intensive, particularly with the increasing size and complexity of data matrices \cite{wang2021robust, song2021approximate, song2022fast}. Addressing this challenge requires approaches that reduce computation time without compromising accuracy.

% \textcolor{red}{[CL: First state the challenges in computation before introducing rSVD.]}

% SVD on texts was originally used for document comparison in Latent Semantic Analysis (LSA) technique introduced by \citet{deerwester1990indexing}. SVD is employed to reduce the number of terms (from document-term matrix) while preserving the similarity between documents.

To compute principal components, we use randomised SVD \cite[rSVD]{halko2011finding} that extracts the column space from unilateral random projections. rSVD utilises randomisation to accelerate the process of finding a low-rank approximation of a matrix. This enables efficient processing of large matrices, significantly reducing computational costs, while also mitigating potential adversarial effects \cite{bingham2001random, xie2017mitigating, taran2019defending}. 
Following \citet{halko2011finding}, we adopt a two-stage framework to approximate a low-rank matrix of a given $m \times n$ matrix $\mathbf{A}$ using randomised algorithms:

% \begin{enumerate}
%     \item \textbf{Step 1}: Compute an approximate basis $\mathbf{Q}$ with $l$ orthonormal columns for the range of $\mathbf{A}$, such that $\mathbf{A} \approx \mathbf{Q} \mathbf{Q}^{\ast} \mathbf{A}$.
%     \item \textbf{Step 2}: Given such a matrix $\mathbf{Q}$, which is much smaller than $\mathbf{A}$, we use it to compute our desired SVD.
% \end{enumerate}

\paragraph{Step 1.} Compute an approximate basis $\mathbf{Q}$ with $l$ orthonormal columns for the range of $\mathbf{A}$, such that $\mathbf{A} \approx \mathbf{Q} \mathbf{Q}^{\ast} \mathbf{A}$.

\paragraph{Step 2.} Given such a matrix $\mathbf{Q}$, which is much smaller than $\mathbf{A}$, we use it to compute our desired SVD.

Motivated by the Johnson-Lindenstrauss lemma \cite{johnson1984extensions}, we explore the preservation of pairwise distances. This lemma demonstrates that such distances among a set of points in a Euclidean space can be approximately maintained when projected into a lower-dimensional Euclidean space. Utilising this principle, we employ random sampling on the range of $\mathbf{A}$. We use a Gaussian random matrix, denoted as $\mathbf{\Omega} \in \mathbb{R}^{d \times r}$, where $r$ is a sampling parameter indicating the number of Gaussian random vectors. The orthonormal basis for these vectors yields the desired basis $\mathbf{Q}$. This scheme is formally presented in \citet{halko2011finding}.

% Motivated by the Johnson-Lindenstrauss lemma \cite{johnson1984extensions}, which demonstrates that pairwise distances among a set of points in a Euclidean space are approximately preserved when projected into a lower-dimensional Euclidean space, we employ random sampling on the range of $\mathbf{A}$ using a Gaussian random matrix denoted as $\mathbf{\Omega} \in \mathbb{R}^{d \times r}$, where $r$ is a sampling parameter indicating the number of Gaussian random vectors. The orthonormal basis for these vectors yields the desired basis $\mathbf{Q}$, \textcolor{red}{and this scheme is formally presented in \citet{halko2011finding}}.

The efficiency of the rSVD algorithm derives from the fact that $\mathbf{B} = \mathbf{Q}^{*} \mathbf{A}$ is relatively smaller in comparison to $\mathbf{A}$, where $^{*}$ represents the conjugate transpose operation \cite{Turnbull1932AnIT}. This efficiency is based on the observation that $\mathbf{A}$ is approximately equal to $\mathbf{A} \approx \mathbf{Q} \mathbf{Q}^{*} \mathbf{A} = \mathbf{Q} (\mathbf{\tilde{U}} \mathbf{\Sigma} \mathbf{V}^{*})$, allowing us to set $\mathbf{U} = \mathbf{Q} \mathbf{\tilde{U}}$ to obtain a low-rank approximation, resulting in $\mathbf{A} \approx \mathbf{U} \mathbf{\Sigma} \mathbf{V}^{*}$. It is important to note the randomness only occurs during the computation of  $\mathbf{Q}$ matrix, while Step 2 in the SVD computation remains deterministic when $\mathbf{Q}$ is given. Following \citet{halko2011finding}, we employ the subspace iteration method to implement the randomised range finder for obtaining  matrix $\mathbf{Q}$. 
% The underlying concept is that power iterations do not alter the singular vectors, but they do reduce the influence of small singular values.
% In the context of adversarial defence, this means that the algorithm is less sensitive to small changes in the input data.

% By controlling the distribution and dimensionality of random projection matrices \cite{vershynin2018high, dong2023adversarial}, we preserve pairwise distances between data points, as described by the Johnson-Lindenstrauss lemma. 
% It ensures that the distance between any two points in the new low-dimensional space is approximately very close to the distance between the same points in the original high-dimensional space.
% Furthermore, the projection is achieved by a simple linear transformation via random projection matrix onto the token embeddings matrix, whose entities are sampled from a pre-defined distribution.

% \subsection{Temporal Aggregation}
% \label{sec:aggregation}

% Since we want the network to be invariant to temporal position, we propose two aggregation variants that are fully connected only along the embedding dimension axis.

% \subsubsection{Temporal Average Pooling (TAP)}
% \label{sec:tap}

\subsection{Sentence-level Representation}
\label{sec:pfsa}

\begin{figure}
    \centering
    \includegraphics[width=0.9\linewidth]{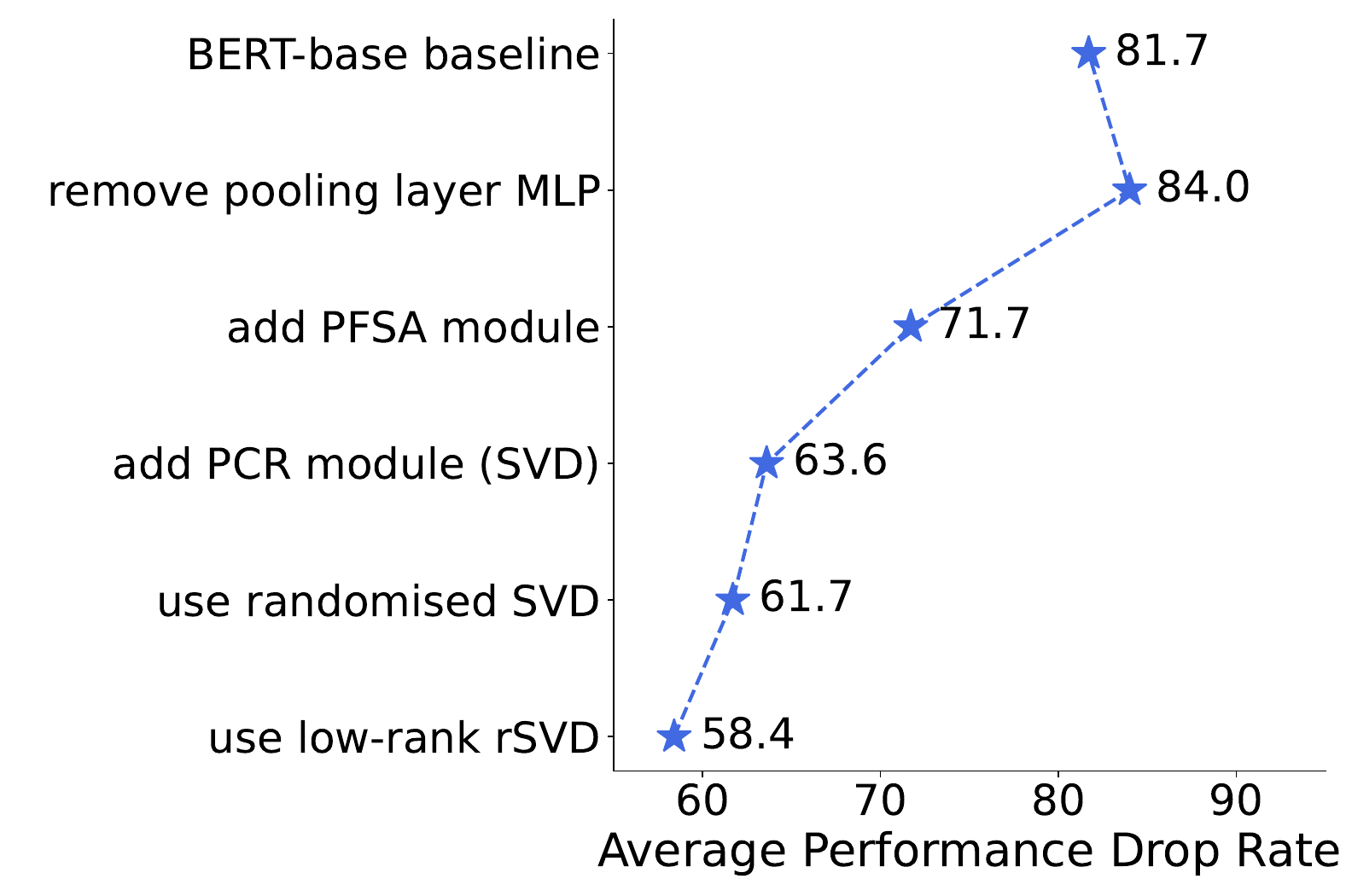}
    \caption{\textbf{The development trajectory} of the module design of \textsc{PuRe}. Each line is based on a modification of the immediately preceding line, tested on the SST2 test set. 
    % A lower Average Performance Drop Rate (\textsc{Apdr}) signifies a more robust model.
    }
    \label{fig:trajectory}
\end{figure}

After obtaining the \textit{purified} token-level representations from the PCR module, we aggregate them to form a single sentence-level representation. 
To do this, we employ Parameter-Free Self-Attention from \citet[PFSA]{zhai2023simple} before the final mean pooling step. PFSA is ideal for this task as it captures global sentence-level features with linear computational complexity and without introducing any trainable parameters. This parameter-free design improves the final semantic representation while mitigating the risk of overfitting. Our ablation study (\S\ref{sec:pcr_pfsa}) confirms that this approach is more effective and efficient than using mean pooling alone.

Finally, Figure.~\ref{fig:trajectory} shows the development trajectory of the module evolved from a standard BERT-base baseline into a model capable of adversarial defence upon integrating the \textsc{PuRe} module. 
More details on the ablation experiments supporting this development trajectory can be found in \S\ref{sec:pcr_pfsa}.

% \subsection{Summary}

% The amalgamation of preliminary concepts (PCA and SVD in \S\ref{sec:preliminaries}) and innovative methodology (PCR and PFSA in \S\ref{sec:pcr} and \S\ref{sec:pfsa}) lays the foundation for our textual adversarial defence approach, positioning us to achieve robust results in our subsequent experiments and analyses.
% To summarise the method, given a tokens embedding acquired from the final hidden layer of a pre-trained model, two modules, PCR and PFSA, compute dominant components removal and complementary attention in a sequential arrangement manner, focusing on "de-noising" and "weighting" respectively. We will discuss experimental results on network engineering in Section~\S\ref{sec:ablation}.
% \red{[CL: this section is pretty redundant. Consider deleting.]}

\section{Experiments}

% \subsection{\textsc{PuRe} Models}
% \label{sec:Pure-models}

% We evaluate two variations of \textsc{PuRe}, each one aggregates different statistics. \textsc{PuRe}$_{\mu}$ aggregates the temporal average vector.  \textsc{PuRe}$_{\mu, \sigma}$ aggregates the temporal average vector and temporal standard deviation vector, and these two vectors are concatenated together and propagated through linear layers and finally the softmax output layer. 

\subsection{Baselines}
\label{sec:baselines}

% We select PGD \cite{madry2018towards}, FreeLB \cite{Zhu2020FreeLB:}, InfoBERT \cite{wang2021infobert}, and TAVAT \cite{li2021token} as our baselines, representing state-of-the-art adversarial training-based methods. For perturbation control-based methods, we include DNE \cite{zhou-etal-2021-defense} and AdvFooler \cite{hoang2024fooling}. Additionally, we incorporate the certification-based approach SAFER \cite{ye-etal-2020-safer}.
% To represent regularisation-based defences, we select Flooding-X \cite{liu2022flooding} and ALS \cite{yang-etal-2023-domain}, which offer structure-free solutions without relying on pre-defined synonym sets.
% For consistency and fair evaluation, we follow the settings outlined in the TextDefender framework \cite{li-etal-2021-searching} across all baselines. 

We use a diverse set of baselines to benchmark \textsc{PuRe}. 
For adversarial training-based methods, we include PGD \cite{madry2018towards}, FreeLB \cite{Zhu2020FreeLB:}, InfoBERT \cite{wang2021infobert}, and TAVAT \cite{li2021token}. 
For perturbation control-based methods, we adopt DNE \cite{zhou-etal-2021-defense} and AdvFooler \cite{hoang2024fooling}. 
For certification-based methods, we adopt SAFER \cite{ye-etal-2020-safer}. 
For regularisation-based methods, we include Flooding-X \cite{liu2022flooding} and ALS \cite{yang-etal-2023-domain}. 
For consistency and fair comparison, all baselines follow the setup outlined in the TextDefender framework \cite{li-etal-2021-searching}.

% We select PGD \cite{madry2018towards}, FreeLB\footnote{FreeLB++ \cite{li2021searching} is excluded due to reproducibility issue as reported at their GitHub repository.} \cite{Zhu2020FreeLB:}, InfoBERT \cite{wang2020infobert}, TAVAT \cite{li2020tavat}, Mixup \cite{si-etal-2021-better}, Manifold Mixup (M-Mixup) \cite{verma2019manifold}, Text Smoothing \cite{wu-etal-2022-text}, and Label Smoothing \cite{yang-etal-2023-domain} as baselines. 

To evaluate the scalability of \textsc{PuRe}, we apply it across a diverse set of model architectures, including encoder-only models such as BERT \cite{devlin2018bert}, RoBERTa \cite{liu2019roberta}, and DeBERTa \cite{he2020deberta, he2021deberta, he2021debertav3}; decoder-only models like OPT \cite{Zhang2022OPTOP} and Qwen2.5 \cite{qwen2, qwen2.5}; and embedding-based models such as BGE \cite{bge_embedding} and GIST \cite{solatorio2024gistembed}. 
All baselines are fine-tuned using their default settings as described in the original papers. 
Further details on the baselines are provided in Appendix~\ref{sec:defence-baselines}, with model architectures, checkpoints, and parameter sizes listed in Appendix~\ref{sec:architecture_full}.

\subsection{Adversarial Attackers for Evaluation}

% Recent research \cite{wang2020defense, du2021combating, wang2023rmlm} aligns with that character-level attacks can be mitigated by grammar detection or spelling checkers while word-level attacks are more imperceptible to humans and harder for deep neural networks to defend against.
% In this sense, we select a range of word-level attackers to evaluate the robustness to adversarial changes. 
We choose three attackers to evaluate the robustness to adversarial changes.
These attacker are leveraged via TextAttack\footnote{\url{https://github.com/QData/TextAttack}} \cite{morris2020textattack} for an extensive comparison between \textsc{PuRe} and the baseline defence strategies. We use default hyperparameters provided by TextAttack library.
% \textcolor{red}{Note that we exclude some adversarial attackers, such as BERT-Attack}~\cite{li2020bert} due to their expensive computation costs\footnote{BERT-Attack takes way more time than other attackers to generate a single adversarial example even when we have selected smaller values for maximum candidates size. This issue is also reported in \url{https://github.com/QData/TextAttack/issues/586}}.

% \paragraph{BAE.} BERT-based adversarial examples (BAE) \cite{garg2020bae} is a technique that modifies the original text by using contextual perturbations from a BERT masked language model (MLM) to generate alternatives for the masked tokens.

% \input{tables/attackers}

\noindent\textbf{TextFooler}~~\cite{jin2020bert} is a black-box adversarial attack method that generates adversarial examples by ranking and replacing important words with semantically and grammatically similar substitutes, aiming to alter model predictions while preserving the fluency of the original text. It demonstrates high attack success rates across NLP tasks like text classification and entailment by using efficient perturbations.

% \noindent\textbf{TextFooler} \cite{jin2020bert}  uses a strategy where the most important and vulnerable words are replaced with synonyms (defined by similarity in the embedding space from \citet{mrkvsic2016counter}) until the classification label changes. 

\noindent\textbf{TextBugger}~~\cite{li2018textbugger} is designed to generate semantic-preserving adversarial texts under both white-box and black-box settings. It uses character- and word-level perturbations to manipulate texts minimally while achieving high attack success rates against real-world NLP applications.

% \noindent\textbf{TextBugger} \cite{li2018textbugger} generates perturbations through a combination of typo-like character edits and synonym substitutions. 

% \noindent\textbf{TextBugger} \cite{li2018textbugger} proposed a multi-level attack framework that uses five bug generation methods:  insert, delete, swap, substitution with visually similar words, and substitution with a semantically similar words.

\noindent\textbf{PWWS}~~\cite{ren2019generating} is a black-box adversarial attack approach. It generates adversarial examples by replacing words based on their saliency and classification probability, ensuring minimal semantic and grammatical disruption while significantly affecting model predictions.

% \noindent\textbf{PWWS} \cite{ren2019generating} takes into account both word saliency and classification probability when selecting substitute words from WordNet. %\cite{miller1995wordnet}.

While we acknowledge the advancements in attack techniques, TextAttack currently provides limited support for newer methods up to 2021. 
Therefore, we focused on three well-established, general-purpose attack methods that are widely recognised for evaluating adversarial robustness \cite{nguyen-minh-luu-2022-textual, wang2022rethinking, wang2022detecting, yang-etal-2023-fantastic, zhan-etal-2023-similarizing, hu-etal-2023-mask, yang-etal-2023-domain, gao-etal-2023-dsrm, shen2023textshield, lu2024less, ji-etal-2024-advancing, zhang-etal-2024-random, zhao2024disentangled}.

% \noindent\textbf{PWWS} \cite{ren2019generating} designed a word-level attack framework that uses WordNet \cite{miller1995wordnet} to build a synonym set that contains all synonyms of a certain word, and incorporates word saliency \cite{li2015visualizing, li2016understanding} into the attack algorithm to determine the word replacement order.

% Table~\ref{tab:attackers} summarises the comparison of the 3 attackers.

\subsection{Evaluation Metrics}
\label{sec:metrics}

Following prior studies \cite{zhan-etal-2023-similarizing, zhao2024disentangled}, we consider four evaluation metrics to measure the resilience of victim models against the aforementioned adversarial attacks. Considering the diverse evaluation metrics across tasks and varying defencing performance across models, we also adopt performance drop rate \cite{zhu2023promptbench} to quantify the relative performance decline.

%\paragraph{\textsc{Acc}.} 

\noindent\textbf{Clean Accuracy (\textsc{Acc})}~~measures the accuracy of the model on the before-attack dataset. It provides a baseline for how well the model performs without adversarial interference.

% \noindent\textbf{\textsc{Acc.}} A higher clean accuracy (\textsc{Acc}) shows the model has better in-distribution generalisation capability.

\noindent\textbf{Accuracy Under Attack (\textsc{Aua})}~~evaluates the accuracy of the model when subjected to adversarial examples. A higher \textsc{Aua} indicates better robustness against adversarial attacks.

% \noindent\textbf{\textsc{Aua}.}~~The accuracy under attack (\textsc{Aua}) evaluates a model's ability to make accurate predictions on adversarial data generated by specific attack methods. A higher \textsc{Aua} indicates a stronger defence against adversarial attacks.

\noindent\textbf{Attack Success Rate (\textsc{Asr})}~~is the percentage of adversarial attacks that successfully cause the model to misclassify. A lower \textsc{Asr} signifies a more robust model.

% \noindent\textbf{\textsc{Asr}.}~~The attack success rate (\textsc{Asr}) \cite{wu2021performance} is calculated by dividing the number of texts that have been successfully altered by a particular attack method by the total number of texts involved. Models that are robust against attacks are expected to have a low \textsc{Asr} score. 
% Given the complementary nature, we get the attack fail rate (\textsc{Afr}) using the equation $\textsc{Asr} + \textsc{Afr} = 1$.

\noindent\textbf{Number of Queries (\textsc{AvgQ})}~~quantifies the average number of queries made to the model by an adversarial attack to achieve success. A higher number implies the model is harder to attack \cite{li2021searching}. 

% \noindent\textbf{\textsc{AvgQ}.}~~The average number of attempts an attacker queries the target model in order to successfully attack a model. If the \textsc{AvgQ} is higher, it means that the model is more difficult to attack \cite{li2021searching}.

\noindent\textbf{Performance Drop Rate (\textsc{Pdr})}~~quantifies the relative performance decline, and provides a normalised measure for comparing different attacks \cite{zhu2023promptbench}. 
\textsc{Apdr} stands for average \textsc{Pdr} across different attacks.
% \textsc{Pdr} is given by: $\textsc{Pdr} = 1 - \frac{\textsc{Aua}}{\textsc{Acc}}$.

% \begin{equation}
%     \textsc{Pdr} = 1 - \frac{\textsc{Aua}}{\textsc{Acc}}
% \end{equation}

% \begin{equation}
%     \textsc{Pdr} = 1 - \frac{\mathcal{M}(f_{\theta}(\mathcal{A}(x)), y)}{\mathcal{M}(f_{\theta}(x), y)}
% \end{equation}
% where $\mathcal{A}(\cdot)$ is the adversarial attack applied to the text inputs, and $\mathcal{M}(\cdot)$ is the evaluation function. For classification task, $\mathcal{M}(\cdot)$ is the indicator function $\mathbbm{1} [\hat{y}, y]$ which equals to 1 when $\hat{y} = y$, and 0 otherwise.

\subsection{Datasets}

We evaluate \textsc{PuRe} across eight language understanding datasets covering various NLP tasks such as: sentiment analysis, subjectivity status classification, paraphrase identification, textual entailment, and commonsense reasoning. 
In contrast to other studies \cite{dong2021towards, bao2021defending, li2021searching, wang2022rethinking, shen2023textshield, hu-etal-2023-mask, zeng2023certified, zhan-etal-2023-similarizing, moon-etal-2023-randomized}, which often restrict their evaluations to a limited selection of test samples from their datasets, we extend our analysis to include the entire test sets for all eight datasets, ensuring a comprehensive assessment. 
This broad evaluation approach contrasts with the common practice in the field, where researchers only utilise a small portion of available test data, which may not fully represent the model’s performance across different scenarios. 
% For a detailed comparative analysis of how our evaluation approach stands in relation to current methods, we refer to the Appendix.

% \footnote{This review meticulously examines recent publications from top conferences and journals - detailing how adversarial defence methods are evaluated across various datasets in the field.} section in this paper. 

\noindent\textbf{SST2}~~\cite{socher2013parsing} is a sentiment classification dataset of movie reviews. 

\noindent\textbf{SUBJ}~~\cite{pang2004sentimental} is a review dataset with sentences labelled as subjective or objective.

\noindent\textbf{CR}~~\cite{hu2004mining} is a sentiment classification dataset of customer reviews.

\noindent\textbf{MR}~~\cite{Pang+Lee:05a} is a dataset containing positive and negative sentences from Rotten Tomatoes movie reviews.

\noindent\textbf{MRPC}~~\cite{dolan2005automatically} is a corpus consisting of sentence pairs collected from news-wire articles. Each pair is labelled if it is a paraphrase or not by human annotators.

\noindent\textbf{SICK}~~\cite{marelli-etal-2014-sick} is a large dataset on compositional meaning, annotated with subject ratings for both relatedness and entailment relation between sentences.

\noindent\textbf{SIQA}~~\cite{sap-etal-2019-social} is a commonsense reasoning dataset where the goal is to choose the most appropriate answer from three options to questions about everyday social situations. 

\noindent\textbf{CSQA}~~\cite{talmor-etal-2019-commonsenseqa} is another multiple-choice question answering dataset that requires different types of commonsense knowledge to predict the correct answers. 

% Stanford Sentiment Treebank v2 (SST2) \cite{socher2013parsing} is a sentiment classification dataset of movie reviews. 
% Subjectivity (SUBJ) \cite{pang2004sentimental} is a review dataset with sentence labelled as subjective or objective. 
% Customer Review (CR) \cite{ding2008holistic} is a customer review data set and each sample is labelled as positive or negative. 
% Movie Reviews (MR) \cite{Pang+Lee:05a} is a dataset of containing positive and negative sentences from Rotten Tomatoes movie reviews. 
% Microsoft Research Paraphrase Corpus (MRPC) \cite{dolan2005automatically} is a corpus consists of sentence pairs collected from newswire articles. Each pair is labelled if it is a paraphrase or not by human annotators.
% Sentences Involving Compositional Knowldedge (SICK) \cite{marelli-etal-2014-sick} is a large dataset on compositional meaning, annotated with subject ratings for both relatedness and entailment relation between sentences.
% If the dataset has no official validation set, we randomly select 10\% of the training set to build validation set. 

Note that the test sets of SIQA and CSQA are not publicly available; we evaluate baselines and \textsc{PuRe} on their validation sets. 
Table.~\ref{tab:statistics} summarises the statistics of the four single text classification datasets, two text pairs classification datasets, and two multiple-choice classification datasets.

\subsection{Implementation Details}

% For every pooling strategies, the dimensions of sentence-level representation are $\mathbb{R}^{d}$ for TAP and \textsc{PuRe}, and $\mathbb{R}^{2d}$ for TSTP.
We take the output vector from the pooling layer and use it to construct a feed-forward neural network. We employ an affine transformation followed by a softmax and cross-entropy for classification.
We fine-tune PLMs using AdamW optimiser \cite{loshchilov2017decoupled} for four epochs. 
% We adopt a grid search strategy to choose the best hyper-parameters (batch sizes $\{8, 16, 32\}$, learning rates $\{ 1e^{-5}, 2e^{-5}, 5e^{-5} \}$) for each task. 

To keep experiments simple and reproducible, we avoid extensive hyper-parameter tuning and instead apply a light grid search over a small set of commonly used values: batch sizes $\{8, 16, 32\}$ and learning rates $\{1\mathrm{e}{-5}, 2\mathrm{e}{-5}, 5\mathrm{e}{-5}\}$. 
For the three adversarial attackers, all the four evaluation metrics are tested on the entire test set for every dataset on sequence classification tasks. 

For commonsense reasoning datasets, we follow \citet{branco-etal-2021-shortcutted}, converting the multiple-choice task into a sequence-ranking problem, as outlined in \citet{liu2019multi}. We process the elements of input pairs separately, generating a score for each, with the maximum score corresponding to the selected answer. 
More training details can be found in our public source code\footnote{\url{https://github.com/PuReDefence/PuRe}}.

\section{Results and Analysis}
\label{sec:results}

This section compares \textsc{PuRe} to other baselines in several configurations across datasets and attacks. 
For simplicity, if not specified, we refer the backbone to BERT-base in the following analysis.
% \footnote{We observe the same general trends across all models, and therefore present the results for BERT-base here and the others in Appendix.~\ref{sec:architecture_full}.}.
 
%\subsection{Main Results}

% \subsection{Robustness to Distorted Texts}
\subsection{Generalisation and Robustness}

\input{tables/main_v3}

Table~\ref{tab:main_v3} presents the experimental results for the BERT-base model with various defence methods. 
We observe the same general trends across all models, and therefore present the results for BERT-base here and the others in Appendix~\ref{sec:architecture_full}. 

\textsc{PuRe} performs on-par with the baselines in terms of before-attack accuracy, indicating a good trade-off between robustness and generalisation. 
This trade-off (i.e., higher after-attack accuracy and slightly lower before-attack accuracy) lies in the role of dominant directions in the representation space. 
High-frequency tokens tend to align with top principal components \cite{arora2017simple}. 
Removing these components (most of which are not useful discriminative features, with only a small fraction lying in the dominant vector space) inevitably leads to a minor decrease in clean accuracy, since some discriminative information is lost. 
While \textsc{PuRe} may cause a slight drop in accuracy on clean data, it typically yields much more resilient decision boundaries and improved robustness to adversarial perturbations. 

We observed a notable variation in the \textsc{Apdr} scores across six datasets when subjected to adversarial attacks. 
Specifically, datasets such as SST2, MR, and MRPC exhibit higher \textsc{Apdr} values (58.35\%, 62.03\%, 78.89\% respectively), suggesting these are more challenging to defend compared to SUBJ, CR, and SICK, which demonstrated lower \textsc{Apdr} values (23.62\%, 51.69\%, 53.85\% respectively). 
This variability in resilience may be attributed to inherent dataset characteristics, including the complexity of the text, the diversity of linguistic expressions, and the nature of the tasks involved. 
For instance, simpler datasets like SST2 might be more susceptible to semantic shifts caused by adversarial perturbations due to their straightforward linguistic structures. 
Conversely, datasets like SICK, involving more complex semantic relationships, might inherently diffuse such attacks more effectively. 
Thus, our subsequent analysis will primarily focus on SST2, MR, and MRPC datasets.

\subsection{Adversarially-augmented Training}
\label{sec:aat}

\begin{figure}
    \centering
    \includegraphics[width=1.0\linewidth]{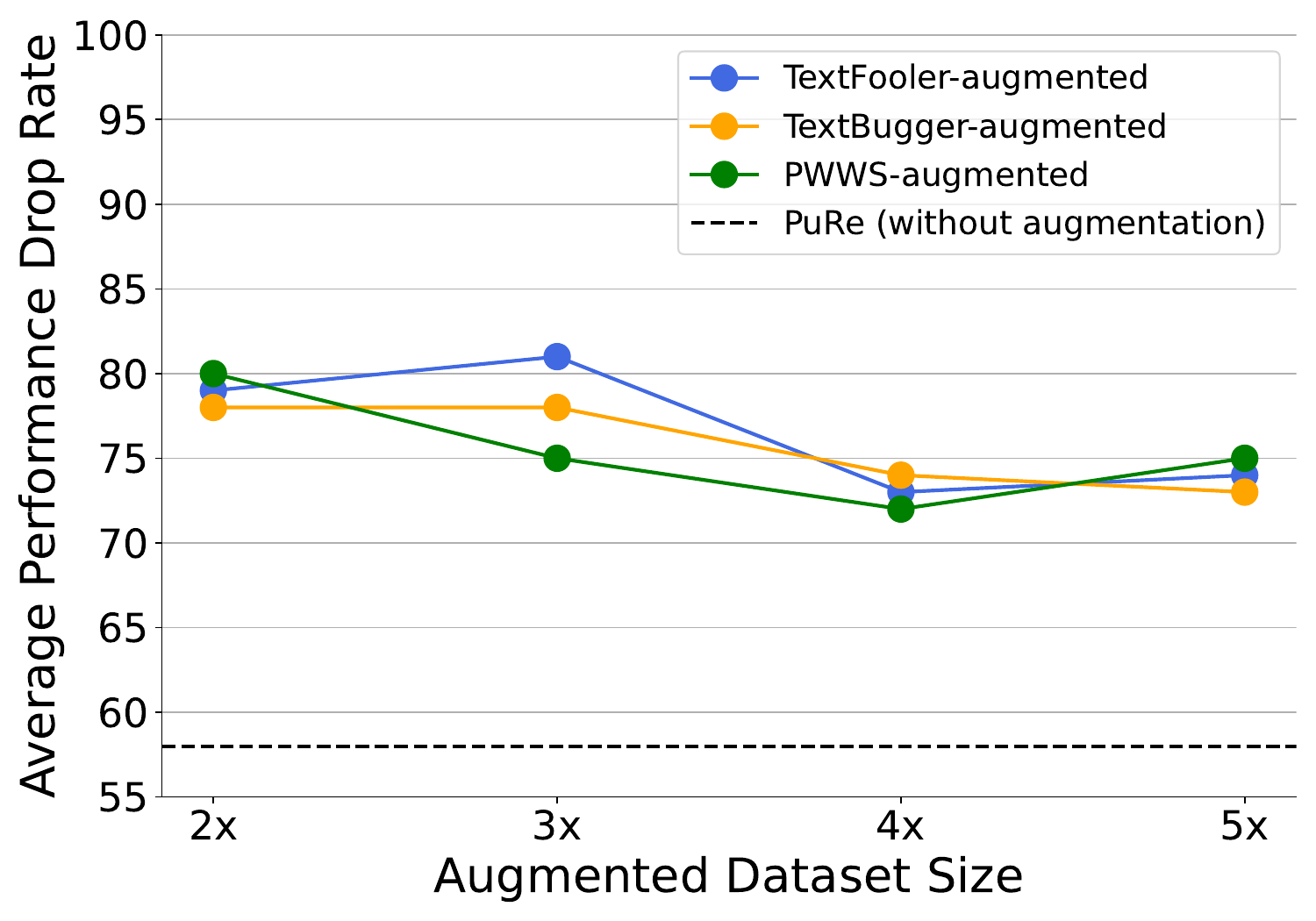}
    \caption{\textsc{Apdr} comparison of the AdvAug training using BERT-base model on the SST2 test set. While AdvAug improves robustness, \textsc{PuRe} achieves a higher \textsc{Apdr} without incurring the computational overhead of generating and incorporating adversarial examples.
    }
    \label{fig:advaug}
\end{figure}

We perform an AdvAug experiment on the SST-2 dataset by augmenting the training set with adversarial examples that preserve the original labels. Each training sample is initially paired with one adversarial counterpart, resulting in a 2x dataset. 
To investigate the effect of larger-scale augmentation, we further expand the dataset by generating up to four distinct adversarial examples per input, creating datasets up to 5x the original size. These augmented datasets are then used to fine-tune a BERT-base model under same training configurations.

As shown in Figure~\ref{fig:advaug}, increasing the size of the augmented dataset generally leads to a decrease in \textsc{Apdr}, indicating improved robustness. 
However, this improvement tends to plateau beyond a certain point, particularly around the 4x and 5x augmentation levels, suggesting diminishing returns from simply scaling up adversarial data. 
Moreover, despite the increased exposure to adversarial examples, none of the AdvAug configurations are able to match the robustness achieved by \textsc{PuRe}, which attains a substantially lower \textsc{Apdr} of 58.35\% without requiring any adversarial examples during training. 
These findings underscore the efficiency and effectiveness of \textsc{PuRe}, which offers strong adversarial robustness without incurring the computational overhead associated with extensive adversarial data generation and augmentation.

% (i.e. generating adversarial examples $n$ times larger is approximately equivalent to $n$-fold increase in training duration).

\subsection{Comparing with Different Models}
\label{sec:architectures}

% \url{https://huggingface.co/sentence-transformers/nli-bert-base}
% \url{https://huggingface.co/roberta-base}

\begin{figure*}
    \centering
    \includegraphics[width=0.85\linewidth]{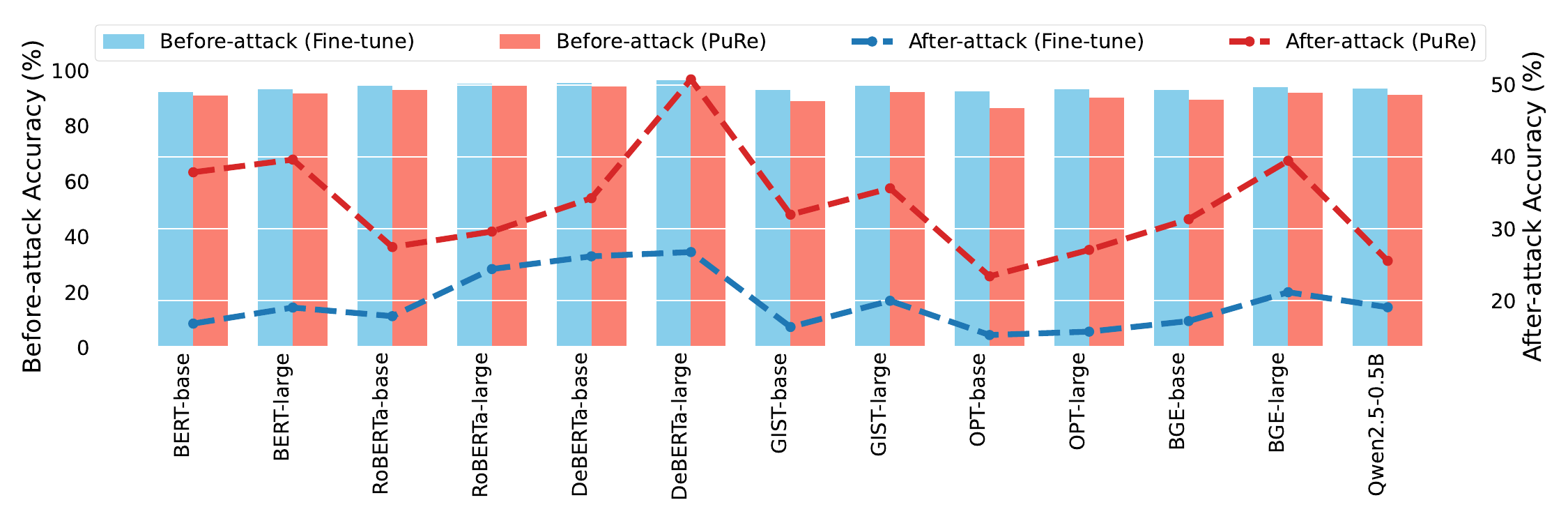}
    \caption{Comparison of before-attack (bar plots) and after-attack (line plots) accuracy on the SST2 test set across various model architectures for both the standard fine-tune baseline and our proposed \textsc{PuRe} approach. The left y-axis shows the models' performance before adversarial attacks, while the right y-axis shows their performance after attacks. \textsc{PuRe} consistently achieves higher after-attack accuracy while maintaining competitive before-attack performance, demonstrating its enhanced adversarial robustness.}
    \label{fig:models}
\end{figure*}

% We compare \textsc{PuRe} with different model architectures: BERT \cite{devlin2018bert}, RoBERTa \cite{liu2019roberta}, DeBERTa-base \cite{he2020deberta, he2021deberta, he2021debertav3}, OPT \cite{Zhang2022OPTOP}, BGE \cite{bge_embedding}, GIST \cite{solatorio2024gistembed}, and Qwen \cite{qwen2, qwen2.5}.
% The interest of this setting is not to compare models directly, but to evaluate the limits and feasibility of \textsc{PuRe} when changing to different model architectures. 
% Detailed performance of each model architecture is available in Appendix.~\ref{sec:architecture_full}.
% Figure.~\ref{fig:models} compares the performance of various models on the SST2 test set under adversarial attacks using standard fine-tuning and \textsc{PuRe}. 
% Overall, \textsc{PuRe} generally improves robustness against attacks as indicated by increased \textsc{Aua} and decreased \textsc{Asr}, though it comes at the cost of a slightly reduced clean accuracy. 
% From base to large models, \textsc{PuRe} generally demonstrates an improvement in adversarial robustness across different architectures. 
% The enhancement in robustness is especially noticeable for DeBERTa, where the transition from DeBERTa-base to DeBERTa-large under \textsc{PuRe} results in the most significant improvement, indicating that DeBERTa benefits the most from scaling up in model size when using \textsc{PuRe} as a defence strategy. 

In this section, we compare \textsc{PuRe} with different model architectures. The focus of this setting is not to compare models directly, but to assess the limits and feasibility of \textsc{PuRe} when changing to different model architectures. 
As seen in Figure.~\ref{fig:models} and Table.~\ref{tab:architectures_full}, \textsc{PuRe} improves adversarial robustness across all architectures, notably providing large performance gain consistently on masked language models (e.g., BERT, DeBERTa). 
We find that \textsc{PuRe} is less effective for more recent generative-based models like Qwen2.5. 
% We also noticed its effectiveness for the latest generative models like Qwen2.5, which involve more complex learning dynamics due to their stronger capabilities, is less strong. 
We conclude that this is attributed to two factors: 
\begin{enumerate*}[label=(\roman*)]
\item larger generative models encode complex feature spaces, with adversarial perturbations spanning multiple principal components, making single-component removal less effective and requiring task- and model-aware mechanisms for optimal balance; and 
% \item masked models, trained to predict specific tokens, produce embeddings with tighter context alignment, potentially localising adversarial features. 
% In contrast, generative models' broader next-token prediction tasks create more diffused representations \cite{timkey-van-schijndel-2021-bark}, making adversarial noise harder to isolate. 
\item masked models are trained by predicting a masked token based on its surrounding context, encouraging the model to focus heavily on local context (the nearby words). 
Any adversarial noise (e.g., small perturbations designed to trick the model) tends to affect only a few specific dimensions of the feature space, making it easier for \textsc{PuRe} to address. 
\end{enumerate*}

This aligns with findings in \citet{timkey-van-schijndel-2021-bark}, which show that encoder-based models, tend to suffer more from representation degeneration, evidenced by the dominance of a single dimension in their embeddings. 
% While \textsc{PuRe} provides significant improvements across both small and large masked models, as well as smaller generative models, extending its robustness to large-scale generative architectures remains an open challenge. 
% Future work could develop adaptive strategies to extend \textsc{PuRe}'s robustness to large-scale generative models as they become increasingly dominant in NLP. 
%Although \textsc{PuRe} offers limited value for Qwen2.5, 
For Qwen2.5, we further conducted more detailed experiments (in \S\ref{sec:remove_top_k_pc_qwqen}) to explore the impact of removing additional principal components beyond the top-1 on adversarial robustness.

% We compare \textsc{PuRe} with different models: BERT-base \cite{devlin2018bert}, RoBERTa-base \cite{liu2019roberta}, and DeBERTa-base \cite{he2020deberta, he2021deberta, he2021debertav3}. The interest of this setting is not to compare models directly, but to evaluate the limits and feasibility of \textsc{PuRe} when changing to different models. 
% The comparison is limited to vanilla fine-tuning and TAVAT, with the latter being identified as the relatively robust method (according to Table.~\ref{tab:main_v3}). 
% The trend in Table.~\ref{tab:architectures_v2} suggests that while more advanced models like RoBERTa and DeBERTa are more capable and resilient in general (as seen from their superior fine-tuning results), \textsc{PuRe} shows diminished effectiveness on these models compared to BERT. This is likely due to the defence strategy's alignment with simpler models and a potential gap in how well \textsc{PuRe} can handle the richer feature spaces and more sophisticated token interactions in RoBERTa and DeBERTa. 
% To ensure that \textsc{PuRe} remains effective on state-of-the-art architectures, it may need to be adapted to take into account the unique vulnerabilities and representational power of these more advanced models. 
% We leave this detailed exploration of \textsc{PuRe}'s performance across varying architectures to future research. 

% \input{tables/architectures_v2}

\subsubsection{Removing Additional Principal Components beyond the Top-1}
\label{sec:remove_top_k_pc_qwqen}

\input{tables/top_k_pc}

% To understand the impact of removing principal components on adversarial robustness, we conducted preliminary experiments on both encoder-based and decoder-based Transformer models. 
% For encoder-based models (e.g. BERT, RoBERTa, and DeBERTa), removing additional principal components beyond the top-1 resulted in a marked decline in before-attack accuracy. 
% This observation aligns with prior findings by \citet{timkey-van-schijndel-2021-bark}, who reported that masked language models often suffer from representation degeneration, with a single dimension dominating their embeddings. 
% This dominance makes the embeddings less isotropic, leading to a loss of expressive power when additional components are removed.

To understand the impact of removing principal components in adversarial robustness, we were motivated by prior findings from \citet{timkey-van-schijndel-2021-bark}, which highlighted a contrast between encoder-based and decoder-based Transformer models in terms of dimensionality dominance. 
It was found in \citet{timkey-van-schijndel-2021-bark} that, top-1 dimension dominates the cosine similarity contribution between random sentence pairs for encoder-based models; while on the other hand, top-3 dimensions contribute more equally to GPT-2, which is a finding that we hypothesise could generalise to more decoder-based models. 
Building on this insight, we conducted experiments to examine whether such patterns hold for latest state-of-the-art decoder models such as Qwen2.5 \cite{qwen2, qwen2.5}. 
Our results confirmed these observations: for encoder-based models (e.g. BERT, RoBERTa, and DeBERTa), removing additional principal components beyond the top-1 caused a marked decline in before-attack accuracy.
Conversely, our experiments (see Table.~\ref{tab:remove_top_k_pc}) with the decoder-based model Qwen2.5 revealed an intriguing behaviour: removing more than the top-1 principal component improved after-attack accuracy, albeit with a slight reduction in before-attack accuracy. 
Removing the top-1 to top-3 components further enhances robustness while maintaining reasonable accuracy (91.16\%). 
However, removing more components (top-4 and top-5) yields minimal robustness gains but a sharp accuracy drop (87.64\%). 
A connection can be drawn from the above findings and \citet{timkey-van-schijndel-2021-bark}: for models that take more dimensions in embeddings to dominate the cosine similarity computation, removing more than one principle components helps bringing an isotropic embedding space, and improved adversarial robustness. 
However, there exists a trade-off between reaching isotropy and losing too many informative components.
% We conjecture that this improvement stems from the inherently more isotropic nature of decoder-based models, which are less dominated by any single dimension in their token representations compared to encoder-based models. 

\subsection{Commonsense Reasoning Task}

Following prior work \cite{branco-etal-2021-shortcutted}, we adopted only TextFooler to evaluate the adversarial performance under same experimental settings. Table.~\ref{tab:reasoning} presents the results of various defence methods on commonsense reasoning datasets, using BERT-base as the underlying architecture. 
We compare \textsc{PuRe} exclusively against regularisation-based defence methods, as these approaches do not rely on prior knowledge of the adversary's synonym generation. 
Overall, \textsc{PuRe} emerges as a strong adversarial defence method in the context of commonsense reasoning tasks, balancing both before-attack performance and robustness to adversarial perturbations. 
These findings offer evidence for further exploration of \textsc{PuRe}'s applicability to a wider range of NLP tasks.

\input{tables/reasoning}

\subsection{Natural Robustness in \textsc{PuRe}}
\begin{figure*}
    \centering
    \includegraphics[width=0.85\textwidth]{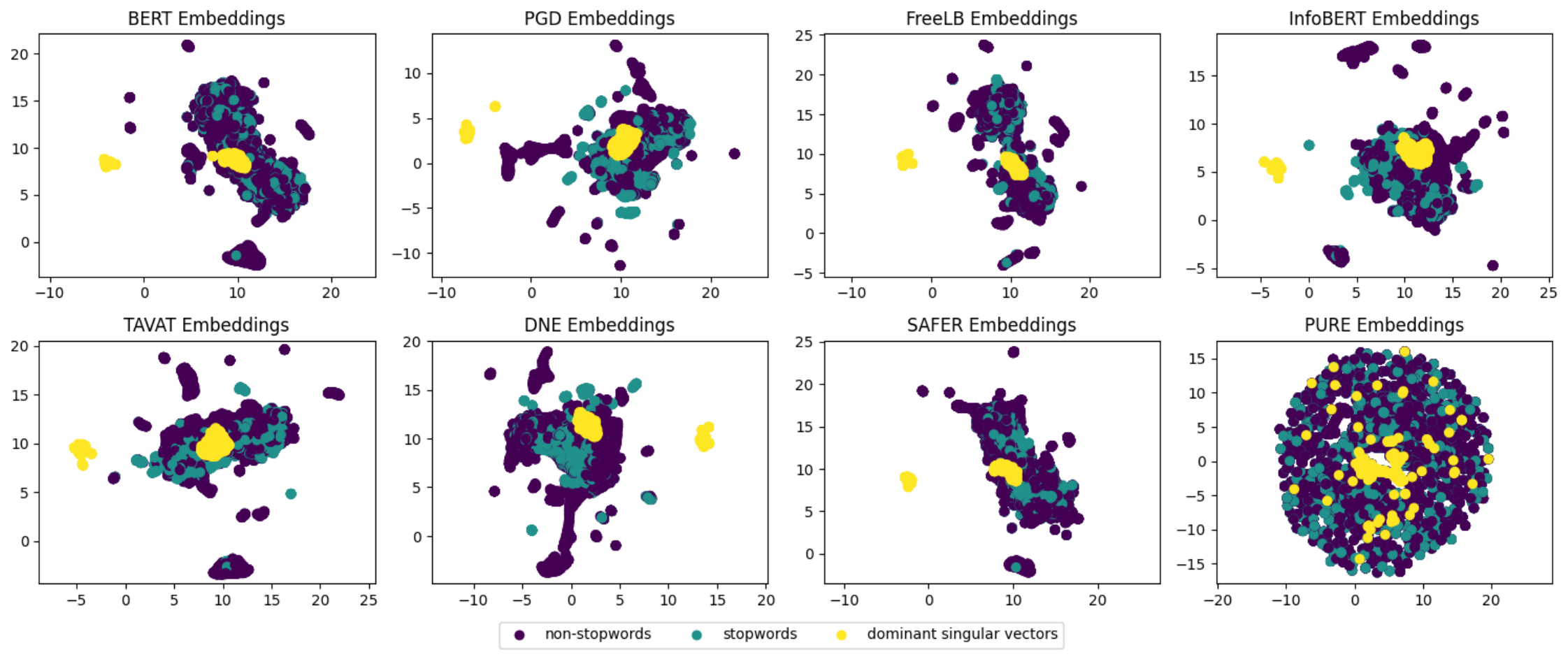}
    % \caption{Embedding spaces of different defence methods for SST2 plotted using UMAP \cite{mcinnes2018umap}.}
    \caption{Each token (not each sentence) is projected onto a 2D subspace using UMAP \cite{mcinnes2018umap}. The baselines exhibit anisotropic distributions: stopword tokens (green) cluster near dominant singular vector directions (yellow), consistent with findings that high-frequency tokens tend to align with top principal components \cite{arora2017simple}. This alignment creates predictable directions that adversaries can exploit. In contrast, \textsc{PuRe} disperses both stopwords and dominant components, resulting in a more isotropic distribution that substantially reduces concentrated adversarial attack surfaces.
    %\red{Each token (not each sentence) is projected onto a 2D subspace via UMAP \cite{mcinnes2018umap}. Baselines exhibit anisotropic distributions: stopwords points (green) are near dominant singular vectors clusters (yellow), aligning with findings that high-frequency tokens correlate with top principal components \cite{arora2017simple}. This creates predictable directions adversaries exploit. \textsc{PuRe} disperses both stopwords and dominant vectors, and the resulting isotropic distribution largely mitigates concentrated adversarial attack surfaces.}
    }
    \label{fig:dsv}
\end{figure*}
\input{tables/intra_sim}

In this section, we illustrate a key property of \textsc{PuRe}: \textit{natural robustness}. 
% This robustness is termed ``natural'' because it is achieved without resorting to iterative min-max optimisation. 
This is termed natural because model's robustness is achieved without \textit{explicit} adversarial defence methods. 
First, we discuss the relationship between robustness and isotropy. 
As depicted in Figure.~\ref{fig:dsv}, \textsc{PuRe} maps each input sentence to a lower-dimensional space, effectively bringing perturbed and normal sentences into closer proximity in a more uniform distribution. 
Then, the adversarial examples are somehow treated as normal samples in the embedding space, smoothing the attack. This means that the perturbed parts in adversarial examples will take a weaker effect on the victim models. 
A parallel can be drawn with the findings in \citet{arora2016latent}, which details that the isotropy has a ``purification'' effect that mitigates the (rather large) approximation error in the PMI models \cite{church1990word}, and underscores the power of high-dimensional geometry to retain structure through isotropic regularisation in embeddings. 
% A parallel can be drawn with the findings in \textsc{Rush} \cite{pang2022rush}, which details the connection between robustness and Lipschitz continuity and demonstrates the influence of contrastive learning on this continuity.

We further investigate the natural robustness of \textsc{PuRe} by assessing intra-sentence similarity scores \cite{xiao2023isotropy}, illustrated in Table~\ref{tab:intra-sim}, revealing the isotropic characteristics of \textsc{PuRe}. 
Specifically, \textsc{PuRe} increases the unadjusted intra-sentence similarity from 0.592 to 0.895, highlighting its effectiveness to induce a more robust and semantically rich sentence-level representation. 
% Based on the results in Table.~\ref{tab:intra-sim} and Figure.~\ref{fig:dsv}, we conjecture that \textsc{PuRe} can benefit from the concentration of measure phenomenon \cite{milman1986asymptotic} in high-dimensional spaces. 
This isotropy property reduces the likelihood of noise dominating any single direction in the latent space, while preserving meaningful semantic structures. 
Isotropy in \textsc{PuRe} can be seen as a high-dimensional analog of the Johnson-Lindenstrauss \cite{johnson1984extensions} property, where the post-\textsc{PuRe} contributes equally across dimensions and maintaining the semantic structure of the data. 

\subsection{PCR and PFSA Modules}
\label{sec:pcr_pfsa}

\input{tables/pcr_pfsa}

In this section, we explore the impact of the PCR and PFSA modules through an ablation study on the SST2 and MRPC datasets, as shown in Table.~\ref{tab:pcr_pfsa}.
While PCR and PFSA are effective individually, significantly surpassing the baselines (see Table.~\ref{tab:main_v3}), their combined use within \textsc{PuRe} leads to substantial improvements (lower \textsc{Pdr} and higher \textsc{Aua}) in resisting adversarial attacks. 

\subsection{Analysis of rSVD in PCR}

\input{tables/rand_v2}

One of the most notable findings is the superior performance of rSVD compared to SVD when integrated into PCR, as demonstrated in Table.~\ref{tab:rand_v2}. It highlights the \textsc{Aua} and \textsc{Pdr} scores of the BERT-base model using both SVD and rSVD on the SST2 and MRPC datasets, clearly showing the superiority of rSVD.

The inherent stochastic nature of rSVD, which involves the introduction of a Gaussian matrix $\Omega$ as described in \S\ref{sec:rsvd}, introduces a level of randomness that serves as implicit regularisation. The randomisation in rSVD potentially enhances the model's robustness. We speculate that this robustness manifests as an increased difficulty for adversaries to craft effective attacks, due to the unpredictable nature of the decomposition's outcome. This aligns with previous studies \cite{moon-etal-2023-randomized, zeng2023certified}, which have shown the benefits of randomness in improving adversarial defences.

\subsection{Run Time Analysis}
\label{sec:runtime}
% Can refer to RSMI paper.
% In Appendix~\ref{appendix: computational cost}, we demonstrated the efficiency of \textsc{PuRe} in computation.

\input{tables/runtime}

We compare the computation speed of \textsc{PuRe} with the baselines on the BERT-base model fine-tuned on MRPC because this dataset has the longest average sequence length. 
All experiments are carried out on a single RTX 4090 GPU. 
Following prior work \cite{wang-lin-2025-tougher}, we adjust the number of gradient computation steps for PGD, FreeLB, and InfoBERT to 5, aligning other parameters with the default configurations as specified in their respective original papers. 
Pre-processing times for DNE and SAFER were excluded to maintain comparability. 
As shown in Table.~\ref{tab:runtime}, while fine-tuning serves as a baseline with a run time of 1.0 for both training and inference, PGD and InfoBERT exhibit significantly higher training costs (x3.3 and x3.8, respectively) despite similar inference times. 
While baselines like Flooding-X and ALS require slightly less runtime than \textsc{PuRe}, their robustness performance is substantially weaker compared to \textsc{PuRe}.
Additionally, \textsc{PuRe} offers a more efficient solution, particularly with the rSVD variant. 
These results (i.e. Table.~\ref{tab:rand_v2} and Table.~\ref{tab:runtime}) indicate that the randomisation in rSVD not only reduces computational costs but also enhances robustness against adversarial attacks, making it a superior choice over standard SVD without any apparent trade-off in accuracy. 

\section{Conclusion}
\label{sec:conclusion}

In this work, we propose a simple yet effective adversarial defence method called \textsc{PuRe}, which has \textit{natural robustness} against adversarial attacks. 
\textsc{PuRe} is designed as an easily integrable add-on module, based on a straightforward variant of PCA, enabling seamless application to off-the-shelf PLMs with minimal modifications.
\textsc{PuRe} was rigorously evaluated across eight diverse language understanding datasets, demonstrating that \textsc{PuRe} not only enhances adversarial defence but also strikes a balance between robustness and generalisation. 
% Our evaluation of \textsc{PuRe}'s robustness is based on widely used frameworks such as TextAttack, focusing on general-purpose attacks common in sequence classification tasks. 
% While these frameworks provide valuable insights, future work could explore newer or more specialised attacks to more thoroughly assess \textsc{PuRe}'s robustness against the evolving landscape of adversarial methods. 
Our evaluation is conducted using the TextAttack framework, focusing on general-purpose attacks relevant to sequence classification and commonsense reasoning tasks. 
While our evaluation provides strong evidence of effectiveness, future work may consider expanding to newer or more specialised attacks to further validate \textsc{PuRe}'s robustness. 
Additionally, although \textsc{PuRe} requires only standard fine-tuning (i.e., without the need for adversarial examples or custom regularisation), it is not entirely training-free; the PLM still requires fine-tuning with \textsc{PuRe} integrated in order to refine the embedding space for the downstream task. 
Nevertheless, the simplicity, effectiveness, and compatibility of \textsc{PuRe} highlight its potential as a foundational component for building robust NLP systems.

% \red{Our evaluation of \textsc{PuRe}'s robustness is based on widely-used frameworks like TextAttack, focusing on general-purpose attacks prevalent in sequence classification tasks. 
% While these frameworks provide valuable insights, future work can consider evaluations on newer or more specialised attacks to more thoroughly assess \textsc{PuRe}'s robustness against the evolving landscape of adversarial methods. 
% Furthermore, while \textsc{PuRe} requires only standard fine-tuning without the need for adversarial examples or custom regularisation, it is not entirely training-free; the PLM still requires fine-tuned with \textsc{PuRe} integrated to refine the embedding space for the downstream task. 
% %Our current experiments are conducted on models under 1B parameters due to resource constraints, and future work should extend our study to larger state-of-the-art models to better understand \textsc{PuRe}'s scalability and potential trade-offs.
% }
% Future work can extend \textsc{PuRe} to explore its adaptability in multi-modal or generative language models, as well as its performance under more sophisticated adversarial settings. 
% Nevertheless, the simplicity, effectiveness, and compatibility of \textsc{PuRe} highlights its potential as a foundational component for advancing robust NLP systems. 

%We release our code at \url{https://github.com/PuReDefence/pure}.

%\vfill\pagebreak

\section*{Limitations}
\label{sec:limitations}

\noindent\textbf{Adversarial Attacks.}~~We assess \textsc{PuRe}'s robustness using TextAttack, a widely used NLP adversarial benchmark that includes methods prior to 2021. 
While sufficient for general-purpose evaluation, it does not cover newer attacks. As adversarial techniques evolve, future work should incorporate broader evaluations. Additionally, adversarial NLP remains limited in realism: most perturbations are lexical and less representative of real-world threats compared to imperceptible manipulations in computer vision \cite{chen-etal-2022-adversarial}.

% \noindent\textbf{Scalability to Larger Models.}~~\red{Our current experiments are conducted on models under 1B parameters (including a 0.5B version of Qwen2.5), primarily due to resource constraints. 
% In practice, 7B and larger models require GPUs with substantially higher memory, and most recent fine-tuning research on such models uses techniques like LoRA \cite{hu2022lora}, which still necessitate large memory and do not allow full fine-tuning. 
% This makes it challenging to conduct ablation studies that isolate the contribution of \textsc{PuRe} from the effects of specialised fine-tuning methods on larger architectures. 
% Consequently, while our results on smaller models are promising, it remains an open question whether \textsc{PuRe} will remain as effective on larger state-of-the-art models. 
% We encourage future work to extend our study to these larger architectures to better understand \textsc{PuRe}'s scalability and potential trade-offs in a rapidly evolving landscape. 
% }

\noindent\textbf{Scalability to Larger Models.}~~Our study is limited to models under 1B parameters due to hardware constraints. Larger models (e.g., 7B+) require significantly more memory and are typically fine-tuned with methods like LoRA \cite{hu2022lora}, complicating clean ablations. It is unclear whether \textsc{PuRe}'s gains extend to such models, and we encourage future work to explore its scalability and compatibility with larger architectures.

% \noindent\textbf{Not Training-free.}~~While \textsc{PuRe} requires only standard fine-tuning without the need for adversarial examples generation or custom regularisation terms, it still involves the regular training time as fine-tuning. 
% \red{Rather than simply adding the \textsc{PuRe} module to an off-the-shelf fine-tuned model, the PLM must be fine-tuned with \textsc{PuRe} integrated. }
% This means that acquiring robustness is not training-free, as fine-tuning remains a necessary step. 
% \red{This fine-tuning is important to refine the embedding space into a more isotropic structure that is specifically tailored to the downstream task distribution. }
% Therefore, \textsc{PuRe} requires access to the model's architecture and gradients during fine-tuning. 
% This makes it unsuitable for black-box scenarios where only input-output queries are accessible, limiting its applicability in certain real-world settings like API-based models. 

\noindent\textbf{Not Training-free.}~~Although \textsc{PuRe} avoids adversarial training or custom regularisers, it still requires fine-tuning with the module integrated. Thus, it is not training-free and assumes access to model gradients, making it unsuitable for black-box or API-only scenarios.

\section*{Ethics and Broader Impact}
\label{sec:ethics}

The inherent nature of adversarial attacks raises ethical concerns, as malicious users may leverage theoretical adversarial attack literature to develop dangerous tools for the misuse of deployed deep learning systems. 
It is crucial to emphasise that the present study diverges from proposing novel adversarial attack techniques. 
Instead, its focus lies in devising a methodology to alleviate the susceptibility of the most vulnerable or adversarial examples within the neural network. 
Consequently, this specific research endeavor does not give rise to perceived ethical concerns.

\section*{Acknowledgment}
\label{sec:acknowledgment}

We are grateful to the action editor and the anonymous reviewers for their thoughtful feedback and valuable suggestions, which have helped improve the quality and clarity of this work. 
Stuart E. Middleton was supported by the Economic and Social Research Council (ES/V011278/1) and Engineering and Physical Sciences Research Council (EP/Y009800/1).

%\vfill\pagebreak

\bibliography{tacl2021}
\bibliographystyle{acl_natbib}

\appendix

\section{Preliminaries: PCA and SVD}
\label{sec:preliminaries}

Principal Component Analysis \cite[PCA]{abdi2010principal} and Singular Value Decomposition \cite[SVD]{golub1971singular} are cornerstone techniques in the field of machine learning. PCA seeks to transform a set of possibly correlated variables into a smaller number of uncorrelated variables called principal components, with the first principal component accounting for the largest possible variance in the data. This transformation is achieved by identifying the eigenvectors of the data covariance matrix, which correspond to the directions of maximum variance. On the other hand, SVD decomposes a matrix $\mathbf{X}$ into three distinct matrices 
\begin{equation}
    \mathbf{U} \mathbf{\Sigma} \mathbf{V^\top} \gets \text{SVD}(\mathbf{X})
\end{equation}
where $\mathbf{U}$ and $\mathbf{V}$ contain the left and right singular vectors, and $\mathbf{\Sigma}$ is a diagonal matrix with singular values. These singular values are crucial as they measure the importance of each corresponding singular vector in capturing the variance of data. 
The expression $\text{SVD}(\mathbf{X})$ can also be rewritten as a sum of the outer products of the singular vectors, weighted by the singular values (i.e. linear combination of rank-1 matrices):
\begin{equation}
    \mathbf{U} \mathbf{\Sigma} \mathbf{V^\top} = \sum\limits_{i=1}^{k} \sigma_i \mathbf{u}_i \mathbf{v}^\top_i
\end{equation}
where $\mathbf{u}_i$ and $\mathbf{v}_i$ are the columns of $\mathbf{U}$ and $\mathbf{V}$ called the left-singular vectors and right-singular vectors, respectively, and $k$ is the rank of the matrix $\mathbf{X}$. Here, each term $\sigma_i \mathbf{u}_i \mathbf{v}^\top_i$ represents a rank-1 matrix, and the sum of these rank-1 matrices approximates the original matrix $\mathbf{X}$.

Both PCA and SVD are intrinsically related. PCA can be performed through SVD by decomposing the data matrix $\mathbf{X}$ and then using the singular vectors as the principal components. The elegance of SVD, beyond dimensionality reduction, lies in its ability to provide a mathematically rigorous and computationally efficient method for identifying the underlying structure of data.
In \S\ref{sec:pcr}, we propose to use PCA and SVD to enhance the adversarial robustness of NLP models.

\section{Datasets}

\input{tables/stats_full}

In this paper, we fine-tune PLMs on eight datasets: SST2 \cite{socher2013parsing}, SUBJ \cite{pang2004sentimental}, CR \cite{hu2004mining}, MR \cite{Pang+Lee:05a}, MRPC \cite{dolan2005automatically}, SICK \cite{marelli-etal-2014-sick}, SIQA \cite{sap-etal-2019-social}, and CSQA \cite{talmor-etal-2019-commonsenseqa}. 
A detailed description of each dataset is available in Table.~\ref{tab:statistics}.

\section{Adversarial Defence Baselines}
\label{sec:defence-baselines}

\noindent\textbf{PGD}~~\cite{madry2018towards} is a simple method to obtain adversarial perturbations. The process of PGD can be represented as the following min-max problem: 

\begin{equation}
    \min_{\theta} \mathbb{E}_{(x, y) \sim \mathcal{D}} \left[ \max_{\delta} \mathcal{L} \left( f_{\theta}(x + \delta), y \right) \right]
\end{equation}
where $(x, y)$ is the data points in the dataset $\mathcal{D}$ ($x$ is the input sample and y is the corresponding label), $f_{\theta}(x)$ is the model with parameters $\theta$, $\delta$ is the perturbation added to $x$, and $\left\Vert \delta \right\Vert \leq \epsilon$ enforces a constraint on the perturbation budget $\epsilon$.

PGD generates the adversarial perturbation $\delta$ iteratively. For each iteration $t$, update $\delta^{(t)}$ by performing a gradient ascent step $\alpha$ to increase the loss.

\begin{equation}
    \delta^{(t+1)} = \delta^{(t)} + \alpha \cdot \text{sgn} \left( \nabla_{\delta} \mathcal{L} \left( f_{\theta}(x + \delta^{(t)}), y \right) \right)
\end{equation}
where $\nabla_{\delta} \mathcal{L} \left( f_{\theta}(x + \delta^{(t)}), y \right)$ is the gradient of the loss with respect to $\delta$, and $\text{sgn}(\cdot)$ takes the sign of each component in the gradient.

\noindent\textbf{FreeLB}~~\cite{Zhu2020FreeLB:} extends PGD by performing multiple mini-batch updates to craft adversarial examples. 
That is, it combines the adversarial perturbation with large-batch optimisation and reuses gradients across multiple steps, effectively increasing efficiency without requiring separate gradient calculations for each step. 
It simultaneously accumulates the ``free'' parameter gradients $\nabla_{\delta} \mathcal{L}$ in each iteration.

\noindent\textbf{SAFER}~~\cite{ye-etal-2020-safer} employs randomised smoothing techniques to certify that a model's prediction remains consistent within a defined radius of perturbation. 

\begin{equation}
    f_{RS}(\textbf{X}) = \arg \max_{c \in Y} P_{\textbf{Z} \sim \Pi_\textbf{X}} \left( f(\textbf{Z}) = c \right)
\end{equation}
where $f_{RS}$ represents the smoothed classifier, $\textbf{X}$ is the original sentence, $c$ is a class in the label set $Y$, $\Pi_\textbf{X}$ denotes the distribution of perturbed sentences around $\textbf{X}$, and $f(\textbf{Z})$ is the classifier's prediction for a perturbed input $\textbf{Z}$. 
SAFER averages predictions over randomly perturbed versions of the input to certify robustness against adversarial word substitutions.

\noindent\textbf{InfoBERT}~~\cite{wang2021infobert} enhances adversarial training by maximising mutual information between clean and adversarial samples, promoting alignment between original and perturbed representations.

\begin{equation}
    \max I(Y; \mathbf{T}) - n \beta \sum_{i=1}^n I(\mathbf{X}_i; \mathbf{T}_i) + \alpha \sum_{j=1}^M I(\mathbf{T}_{k_j}; \mathbf{Z})
\end{equation}
where $I(Y; \mathbf{T})$ is the mutual information between label set $Y$ and the learned representation $\mathbf{T}$, to retain task-relevant information. 
$n \beta \sum_{i=1}^n I(\mathbf{X}_i; \mathbf{T}_i)$ is the information bottleneck regulariser, which minimises mutual information between the input $\mathbf{X}$ and its representation $\mathbf{T}$, removing irrelevant or noisy information that could be vulnerable to adversarial attacks. 
Here, $\mathbf{X}_i$ is the word token, and $\mathbf{T}_i$ is its local feature representation. 
$\alpha \sum_{j=1}^M I(\mathbf{T}_{k_j}; \mathbf{Z})$ is the anchored feature regulariser, which maximises mutual information between selected robust local features $\mathbf{T}_{k_j}$ and the global sentence representation $\mathbf{Z}$, aligning stable local features with the global representation.

\noindent\textbf{TAVAT}~~\cite{li2021token} constructs fine-grained virtual adversarial examples by applying perturbations selectively at the token level instead of a rigid normalisation ball over the entire sequence. 
TAVAT can be summarised by the following objective function, which combines instance-level and token-level perturbations:

\begin{equation}
    \min_{\theta} \mathbb{E}_{(\mathbf{X}, y)} \left[ \max_{\|\delta\| \leq \epsilon} \max_{\|\eta_i\| \leq n_i \cdot \epsilon} L(f_\theta(\mathbf{X} + \delta + \eta), y) \right]
\end{equation}
Here, $\delta$ represents the instance-level perturbation constrained by $\epsilon$, and $\eta_i$ denotes token-level perturbations scaled by $n_i$ for flexibility based on each token's importance. 

\noindent\textbf{DNE}~~\cite{zhou-etal-2021-defense} employs a neighborhood exploration technique to create virtual sentences by mixing the embedding of the original word in the input sentence with its synonyms. 
It can be summarised with the following training objective, using virtual examples sampled from the convex hull of a word and its synonyms:

\begin{equation}
    \min_{\theta} \mathbb{E}_{(\mathbf{X}, y) \sim D} \left[ \max_{\beta} L(f_\theta(\mathbf{X}_{\beta}), y) \right]
\end{equation}
where $\mathbf{X}_{\beta} = \sum_{x_j \in S(x_i)} \beta_j x_j$ represents virtual samples in the convex hull spanned by each word $x_i$ and its synonyms $S(x_i)$, with the weights $\beta$ drawn from a Dirichlet distribution.

\noindent\textbf{Flooding-X}~~\cite{liu2022flooding} is an efficient and computational-friendly algorithm for
improving PLM's generalisation and resistance to adversarial attacks.
They theoretically prove that the vanilla Flooding method \cite{10.5555/3524938.3525366} is able to boost model's adversarial robustness by leading it into a smooth parameter landscape. 
If the original learning objective is $J$, then the modified learning objective $\widetilde{J}$ with flooding is

\begin{equation}
\label{eq:flooding1}
    \widetilde{J}(\theta) = |J(\theta) - b| + b,
\end{equation}
where $b>0$ is the flood level and $\theta$ is the model parameter.

\noindent\textbf{ALS}~~\cite{yang-etal-2023-domain} applies label smoothing during adversarial training, softening the model's predictions and reducing sensitivity to adversarial inputs, particularly in out-of-domain contexts. 
ALS arises from the worst possible smooth label for each input example.

\noindent\textbf{AdvFooler}~~\cite{hoang2024fooling} randomises the latent representation of the input at test time to fool the adversary throughout the attack, which typically involves iteratively sampling of discrete perturbations to generate an adversarial sample. 
It can be mathematically summarised as follows:

\begin{equation}
    \quad z_{i+1} = h_l(z_i + \epsilon)
\end{equation}
where $z_i$ represents the latent representation of the input at the $i$-th layer, $h_l$ denotes the model's $l$-th layer function, and $\epsilon \sim \mathcal{N}(0, \nu \textbf{I})$ is a Gaussian noise with variance $\nu$ added to the latent space at each layer to randomise the representation, thereby confusing adversarial attacks.

\section{Different Model Architectures}
\label{sec:architecture_full}

\input{tables/model_card}

We apply the baselines over a diverse set of model architectures, including encoder-only models such as BERT \cite{devlin2018bert}, RoBERTa \cite{liu2019roberta}, and DeBERTa \cite{he2020deberta, he2021deberta, he2021debertav3}; decoder-only models like OPT \cite{Zhang2022OPTOP} and Qwen2.5 \cite{qwen2, qwen2.5}; and embedding-based models such as BGE \cite{bge_embedding} and GIST \cite{solatorio2024gistembed} to observe the scalability of \textsc{PuRe}. 
The baselines are fine-tuned according to their default configurations presented in the respective papers. 
Table.~\ref{tab:model_card} describes the checkpoints available on the HuggingFace Hub\footnote{\url{https://huggingface.co/models}}. 
Table.~\ref{tab:architectures_full} is the comparison of different model architectures on the SST2 test set.

\input{tables/architecture_full}

\subsection{Out-of-domain Transferability}

% \input{tables/transferability}

% \textsc{PuRe} shows competitive performance, even in challenging scenarios, and generalises effectively across datasets. We conduct transferability experiments on the SST2 (movie reviews), MR (movie reviews), and CR (customer reviews) datasets, the results of which are presented in Table.~\ref{tab:transferability}. These experiments reveal that the \textsc{PuRe}-enhanced BERT-base model, when trained on one dataset, can effectively transfer its robustness to another dataset. Specifically, \textsc{PuRe} can transfer its defence capabilities from SST2 to CR dataset and vice versa. 
% % This evidence underscores significant potential for transferability of \textsc{PuRe}. 

% \section{Transferability}
% \label{sec:transferability_full}

\input{tables/transferability_full}

In Table.~\ref{tab:transferability_full}, \textsc{PuRe} generally shows a competitive \textsc{Apdr} compared to other defence methods when tested in cross-dataset transfer settings, indicating that \textsc{PuRe} adapts well to new domains without sacrificing significant performance, highlighting its transferability. 

\citet{liu2022flooding} proposed that ``flooding'' the loss function by maintaining it near a predefined constant could theoretically improve adversarial robustness by reducing sensitivity to small perturbations. 
However, our empirical findings indicate that Flooding-X fails to enhance adversarial robustness as anticipated, performing poorly across all robustness metrics in transfer scenarios. 
This may stem from differences in experimental setups, datasets, or attack methods, or suggest that Flooding's impact on robustness is more limited than initially claimed. 
Consistent with \citet{zhu-rao-2023-exploring}, we find that Flooding alone does not effectively promote adversarial robustness, despite its potential benefits for reducing overfitting or improving generalisation in benign settings.

We observe that SAFER demonstrates good adversarial robustness specifically in transfer settings, likely due to its unique combination of embedding stabilisation and randomised smoothing techniques. 
Embedding stabilisation reduces the model's sensitivity to small perturbations by replacing words with synonyms or perturbing embeddings, which lessens the impact of attacks relying on fine-grained modifications. 
Furthermore, randomised smoothing adds noise to the embedding space, making the model's outputs less predictable and increasing the query cost for attackers. 
This approach allows SAFER to effectively generalise to new datasets in transfer scenarios, as its robust representations extend well beyond the original training data, offering enhanced protection without requiring adversarial training tailored to each attack type.

The performance of \textsc{PuRe} is compared against various baseline methods across multiple adversarial robustness metrics. 
Although \textsc{PuRe} does not achieve the highest performance across all individual metrics, it demonstrates competitive results, achieving either the best or second-best \textsc{Apdr} outcomes. 
This suggests that \textsc{PuRe} provides a robust balance in terms of adversarial resilience, making it an effective approach for general-purpose robustness (Table.~\ref{tab:main_v3}), even when evaluated on transfer datasets (Table.~\ref{tab:transferability_full}).

\section{PFSA Hyperparameter Analysis}

\input{tables/ablation}

We performed a hyperparameter sensitivity analysis for the PFSA module to identify optimal values for the sampling parameter $r$ and scaling factor $\alpha$, as detailed in Table~\ref{tab:ablation}. 
$r = 8$ and $\alpha = 1.5$ were adopted as the optimal hyperparameters for the PFSA module on the SST2 dataset.

\section{Further Discussion on the Limitations and Scope of \textsc{PuRe}}

While the main body of the paper outlines the primary limitations, this section provides a further discussion on the boundary conditions and potential challenges of \textsc{PuRe}. 
These points represent important directions for future investigation.

\subsection{The Nature of the Top Principal Component}

\textsc{PuRe}'s core hypothesis is that the top principal component (PC) of instance-level embeddings primarily capture common, non-discriminative information (e.g., syntactic patterns, high-frequency word effects) that adversaries exploit. While our results strongly support this for the tasks and models tested, this assumption may not hold universally.

\begin{itemize}
    \item \textbf{Task-Dependent Information:} For certain complex tasks, the top PC might encode discriminative semantic information. For example, in a legal text classification task, a dominant component might represent a key legal concept. In such cases, removing it could harm clean accuracy more significantly than observed in our experiments.
    \item \textbf{Characterising PCs:} Future work could focus on methods to automatically characterise the information contained within the top PCs for a given task before deciding to remove it. This could lead to an adaptive version of \textsc{PuRe} that only applies the removal when the top PCs is identified as ``noise'' rather than ``signal.''
\end{itemize}

\subsection{Task- and Model-Specificity of Optimal PCR}

Our default implementation removes the top-1 PC, which proved highly effective for encoder-based models. However, as shown in our ablation study with Qwen2.5 (\S\ref{sec:remove_top_k_pc_qwqen}), decoder-based models may benefit from removing additional components (e.g., top-3). This highlights that the optimal number of components to remove is likely not a universal constant but depends on:

\begin{itemize}
    \item \textbf{Model Architecture:} Decoder-only and encoder-decoder architectures may distribute information across their embedding dimensions differently than encoder-only models, leading to different anisotropy patterns.
    \item \textbf{Task Complexity:} Simpler tasks like sentiment analysis might have a single, highly dominant ``noise'' component, whereas more complex reasoning tasks might have vulnerabilities distributed across several top components.
\end{itemize}

A ``one-size-fits-all'' approach may therefore be suboptimal. A more advanced implementation of \textsc{PuRe} could involve a mechanism to dynamically determine the optimal number of PCs to remove based on the model and task.

\subsection{Potential Vulnerability to Adaptive Adversaries
}

Our evaluation uses established, general-purpose attackers. A more sophisticated, adaptive adversary who is aware of the \textsc{PuRe} defence mechanism could potentially circumvent it. Such an adversary could formulate a new attack by solving an optimisation problem with an added constraint: the resulting perturbation must lie in a subspace orthogonal to the top PCs that \textsc{PuRe} removes. 

While this would be a significantly harder attack to craft -- especially with the added randomness from rSVD -- it is theoretically possible. Validating \textsc{PuRe} against such adaptive, white-box attacks would be a critical next step to fully assess its robustness in worst-case scenarios.

\subsection{Applicability to Generative and Open-Ended Tasks}

This work focuses on discriminative NLP tasks (e.g., text classification, natural language inference, commonsense reasoning). The applicability of \textsc{PuRe} to generative tasks like text summarisation, machine translation, or dialogue systems remains an open question. 

\begin{itemize}
    \item \textbf{Information vs. Fluency Trade-off:} Generative models rely on the richness of the representation space to produce diverse, fluent, and coherent text. The information removal inherent in \textsc{PuRe}, while beneficial for robustness in classification, might inadvertently ``flatten'' the representation space, leading to more generic or stylistically bland text generation. The very components that \textsc{PuRe} removes might be responsible for encoding subtle details crucial for high-quality generation.
    \item \textbf{Evaluation Challenges:} Evaluating the impact on generation quality is also more complex than measuring accuracy and requires a different set of metrics in an adversarial setting.
\end{itemize}

Future research should explore whether \textsc{PuRe} can be adapted for generative tasks, perhaps by applying it more selectively or with a lower intensity, to strike a balance between robustness and generation quality. 
Further, it will be interesting to theoretically understand how \textsc{PuRe} provides the implicit robustness to text adversarial attacks and mitigates over-confident predictions on the adversarially attacked examples.

\iftaclpubformat

\onecolumn

\appendix

% \section{Author/Affiliation Options as set forth by MIT Press}
% \label{sec:authorformatting}

% Option 1. Author’s address is underneath each name, centered.

% \begin{quote}\centering
%   \begin{tabular}{c}
%     \textbf{First Author} \\
%     First Affiliation \\
%     First Address 1 \\
%     First Address 2 \\
%     \texttt{first.email@example.com}
%   \end{tabular}
%   \ 
%   \begin{tabular}{c}
%     \textbf{Second Author} \\
%     Second Affiliation \\
%     Second Address 1 \\
%     Second Address 2 \\
%     \texttt{second.email@example.com}
%   \end{tabular}

%   \begin{tabular}{c}
%     \textbf{Third Author} \\
%     Third Affiliation \\
%     Third Address 1 \\
%     Third Address 2 \\
%     \texttt{third.email@example.com}
%   \end{tabular}
% \end{quote}

% Option 2. Author’s address is linked with superscript characters to its name,
% author names are grouped, centered.

% \begin{quote}\centering
%     \textbf{First Author$^\diamond$} \quad \textbf{Second Author$^\dagger$} \quad
%     \textbf{Third Author$^\ddagger$}
%     \\ \ \\
%     $^\diamond$First Affiliation \\
%     First Address 1 \\
%     First Address 2 \\
%     \texttt{first.email@example.com}
%      \\ \ \\
%      $^\dagger$Second Affiliation \\
%     Second Address 1 \\
%     Second Address 2 \\
%     \texttt{second.email@example.com}
%      \\ \ \\
%     $^\ddagger$Third Affiliation \\
%     Third Address 1 \\
%     Third Address 2 \\
%     \texttt{third.email@example.com}
% \end{quote}

\fi

\end{document}

%% file: tables/main_v3.tex
\begin{table*}[!htp]\centering
\resizebox{\textwidth}{!}{
\scriptsize
\begin{tabular}{lcrrrrrrrrrrrr}\toprule
\multirow{2}{*}{\textbf{Dataset}} &\multirow{2}{*}{\textbf{Method}} &\multirow{2}{*}{\textbf{\textsc{Acc}}$\uparrow$} &\multicolumn{3}{c}{\textbf{TextFooler}} &\multicolumn{3}{c}{\textbf{TextBugger}} &\multicolumn{3}{c}{\textbf{PWWS}} &\multirow{2}{*}{\textbf{\textsc{Apdr}}$\downarrow$} \\\cmidrule{4-12}
& & &\textbf{\textsc{Aua}}$\uparrow$ &\textbf{\textsc{Asr}}$\downarrow$ &\textbf{\textsc{AvgQ}}$\uparrow$ &\textbf{\textsc{Aua}}$\uparrow$ &\textbf{\textsc{Asr}}$\downarrow$ &\textbf{\textsc{AvgQ}}$\uparrow$ &\textbf{\textsc{Aua}}$\uparrow$ &\textbf{\textsc{Asr}}$\downarrow$ &\textbf{\textsc{AvgQ}}$\uparrow$ & \\\midrule

\multirow{11}{*}{SST2} &Fine-tune &\underline{92.09} &6.32 &93.14 &87.07 &30.37 &67.02 &41.19 &13.78 &85.03 &127.57 &81.73 \\
&PGD &91.21 &12.63 &86.15 &108.44 &36.57 &59.90 &43.99 &22.30 &75.56 &136.51 &73.87 \\
&FreeLB &\textbf{92.15} &10.76 &88.32 &107.30 &36.63 &60.25 &44.32 &20.92 &77.29 &136.10 &75.29 \\
&InfoBERT &91.93 &8.29 &90.98 &98.53 &31.41 &65.83 &42.10 &19.00 &79.33 &133.55 &78.72 \\
&TAVAT &90.50 &12.19 &86.53 &111.46 &36.13 &60.07 &43.73 &\underline{24.00} &\underline{73.48} &\underline{137.93} &73.36 \\
&DNE &86.72 &10.94 &87.38 &105.50 &24.11 &72.20 &\underline{46.16} &19.67 &77.28 &104.85 &78.97 \\
&SAFER &91.76 &7.58 &91.74 &92.96 &32.62 &64.45 &41.05 &14.06 &84.68 &128.81 &80.29 \\
&Flooding-X &91.65 &4.12 &95.51 &81.59 &27.62 &69.86 &39.50 &12.63 &86.22 &127.17 &83.86 \\
&ALS &91.32 &\underline{18.07} &\underline{80.22} &\underline{117.18} &\underline{42.39} &\underline{53.58} &45.42 &22.19 &75.71 &133.40 &\underline{69.83} \\
&AdvFooler &90.55 &11.56 &87.23 &100.98 &39.09 &56.83 &42.65 &17.43 &80.75 &130.73 &74.94 \\
% \cdashlinelr{2-13}
&\textbf{\textsc{PuRe} (Ours)} &90.88 &\textbf{30.37} &\textbf{66.59} &\textbf{134.01} &\textbf{47.17} &\textbf{48.10} &\textbf{58.52} &\textbf{36.02} &\textbf{60.36} &\textbf{139.97} &\textbf{58.35} \\

\midrule
\multirow{11}{*}{SUBJ} &Fine-tune &97.40 &25.30 &74.02 &189.74 &63.15 &35.16 &69.19 &41.95 &56.93 &196.06 &55.37 \\
&PGD &97.15 &46.90 &51.72 &230.91 &77.40 &20.33 &72.48 &60.70 &37.52 &208.30 &36.52 \\
&FreeLB &\textbf{97.50} &44.40 &54.46 &226.76 &76.95 &21.08 &73.76 &58.40 &40.10 &206.93 &38.55 \\
&InfoBERT &97.41 &35.65 &63.40 &207.53 &72.25 &25.82 &72.93 &51.20 &47.43 &201.44 &45.56 \\
&TAVAT &97.05 &\underline{50.25} &\underline{48.22} &\underline{233.11} &\underline{78.45} &\underline{19.17} &72.80 &\underline{62.60} &35.50 &\underline{208.90} &\underline{34.30} \\
&DNE &95.80 &48.45 &49.29 &223.59 &59.45 &37.94 &\underline{84.94} &62.40 &\underline{34.76} &136.63 &40.74 \\
&SAFER &97.25 &32.90 &66.17 &203.86 &68.05 &30.03 &73.23 &46.80 &51.88 &199.27 &49.36 \\
&Flooding-X &97.15 &24.20 &75.09 &189.59 &66.35 &31.70 &72.12 &40.40 &58.41 &195.65 &55.07 \\
&ALS &\underline{97.45} &38.60 &60.39 &217.69 &71.80 &26.32 &73.80 &51.90 &46.74 &202.45 &44.48 \\
&AdvFooler &96.97 &35.96 &62.92 &204.65 &70.19 &27.62 &76.67 &49.11 &49.36 &204.99 &46.63 \\
% \cdashlinelr{2-13}
&\textbf{\textsc{PuRe} (Ours)} &96.75 &\textbf{67.85} &\textbf{29.87} &\textbf{250.60} &\textbf{80.05} &\textbf{17.26} &\textbf{96.61} &\textbf{73.80} &\textbf{23.72} &\textbf{210.42} &\textbf{23.62} \\

\midrule
\multirow{11}{*}{CR} &Fine-tune &92.28 &3.99 &95.68 &81.16 &36.70 &60.23 &35.10 &10.64 &88.47 &127.97 &81.46 \\
&PGD &91.76 &14.63 &84.06 &113.88 &54.52 &40.58 &41.00 &21.28 &76.81 &142.28 &67.15 \\
&FreeLB &92.82 &11.70 &87.39 &103.70 &53.99 &41.83 &41.33 &17.82 &80.80 &136.81 &70.01 \\
&InfoBERT &\textbf{94.15} &10.11 &89.27 &98.24 &48.40 &48.59 &38.91 &15.69 &83.33 &134.16 &73.73 \\
&TAVAT &91.22 &\underline{15.16} &\underline{83.38} &115.63 &\underline{57.18} &\underline{37.32} &41.80 &\underline{25.00} &\underline{72.59} &\underline{143.23} &\underline{64.43} \\
&DNE &88.74 &11.81 &86.69 &\underline{116.49} &35.16 &59.37 &\textbf{52.12} &18.13 &79.50 &111.51 &75.55 \\
&SAFER &\underline{93.09} &9.84 &89.43 &94.43 &44.95 &51.71 &38.23 &13.30 &85.71 &130.17 &75.62 \\
&Flooding-X &91.22 &3.19 &96.50 &84.81 &44.68 &51.02 &36.43 &10.11 &88.92 &132.35 &78.81 \\
&ALS &91.22 &10.11 &88.92 &99.68 &46.81 &48.69 &39.59 &11.44 &87.46 &128.68 &75.02 \\
&AdvFooler &89.64 &11.91 &86.71 &101.37 &48.65 &45.73 &40.33 &15.13 &83.12 &135.76 &71.85 \\
% \cdashlinelr{2-13}
&\textbf{\textsc{PuRe} (Ours)} &88.82 &\textbf{37.23} &\textbf{58.08} &\textbf{138.43} &\textbf{57.98} &\textbf{34.73} &\underline{46.62} &\textbf{33.51} &\textbf{62.28} &\textbf{143.87} &\textbf{51.69} \\

\midrule
\multirow{11}{*}{MR} &Fine-tune &85.64 &5.35 &93.76 &91.59 &24.86 &70.97 &44.25 &13.23 &84.56 &138.90 &83.09 \\
&PGD &85.18 &11.35 &86.67 &122.64 &\underline{36.59} &\underline{57.05} &49.31 &22.42 &73.68 &149.77 &72.47 \\
&FreeLB &\underline{86.30} &7.32 &91.52 &109.88 &30.11 &65.11 &47.50 &17.45 &79.78 &144.52 &78.80 \\
&InfoBERT &\textbf{86.59} &8.26 &90.47 &111.43 &32.27 &62.73 &47.12 &18.76 &78.33 &146.14 &77.18 \\
&TAVAT &84.90 &11.82 &86.08 &\underline{123.43} &34.62 &59.23 &\underline{50.28} &\underline{23.55} &\underline{72.27} &\textbf{151.25} &72.52 \\
&DNE &82.49 &7.04 &91.46 &94.56 &14.67 &82.22 &48.64 &15.70 &80.86 &114.13 &84.88 \\
&SAFER &\underline{86.30} &10.79 &87.50 &105.78 &31.80 &63.15 &47.23 &17.35 &79.89 &140.78 &76.85 \\
&Flooding-X &85.83 &3.47 &95.96 &89.38 &26.17 &69.51 &44.06 &11.26 &86.89 &137.13 &84.12 \\
&ALS &85.65 &\underline{15.38} &\underline{82.04} &116.72 &34.99 &59.15 &50.06 &21.20 &75.25 &142.89 &\underline{72.15} \\
&AdvFooler &83.28 &14.94 &82.06 &106.91 &33.18 &60.16 &50.19 &20.87 &74.94 &144.76 &72.39 \\
% \cdashlinelr{2-13}
&\textbf{\textsc{PuRe} (Ours)} &85.64 &\textbf{25.89} &\textbf{69.74} &\textbf{135.57} &\textbf{40.06} &\textbf{53.18} &\textbf{58.87} &\textbf{31.61} &\textbf{63.05} &\underline{151.08} &\textbf{62.03} \\

\midrule
\multirow{11}{*}{MRPC} &Fine-tune &84.40 &2.32 &97.25 &124.00 &3.25 &96.15 &72.84 &4.41 &94.78 &250.53 &96.06 \\
&PGD &84.06 &9.86 &88.28 &205.38 &11.25 &86.62 &101.98 &16.12 &80.83 &282.70 &85.24 \\
&FreeLB &\underline{85.45} &11.48 &86.57 &212.41 &\underline{11.65} &\underline{86.36} &\underline{107.19} &17.91 &79.04 &\underline{283.42} &83.99 \\
&InfoBERT &\textbf{85.91} &5.22 &93.93 &168.33 &6.72 &92.17 &89.11 &9.91 &88.46 &269.48 &91.52 \\
&TAVAT &84.29 &8.70 &89.68 &\textbf{229.16} &10.43 &87.62 &106.60 &17.22 &79.57 &\textbf{289.42} &85.63 \\
&DNE &73.04 &\textbf{21.97} &\textbf{69.92} &186.46 &4.70 &93.58 &82.30 &\textbf{19.19} &\textbf{74.14} &227.71 &\underline{79.07} \\
&SAFER &84.46 &3.07 &96.36 &121.40 &3.30 &96.09 &70.37 &4.75 &94.37 &249.26 &95.61 \\
&Flooding-X &82.03 &5.04 &93.85 &141.62 &5.51 &93.29 &81.77 &8.52 &89.61 &260.67 &92.25 \\
&ALS &83.77 &4.06 &95.16 &149.17 &6.03 &92.80 &82.89 &9.10 &89.13 &260.25 &92.36 \\
&AdvFooler &83.46 &4.67 &94.40 &150.98 &7.64 &90.85 &90.13 &6.84 &91.80 &267.37 &92.35 \\
% \cdashlinelr{2-13}
&\textbf{\textsc{PuRe} (Ours)} &82.20 &\underline{17.22} &\underline{79.07} &\underline{226.10} &\textbf{16.29} &\textbf{80.20} &\textbf{107.73} &\underline{18.55} &\underline{77.45} &273.21 &\textbf{78.89} \\

\midrule
\multirow{11}{*}{SICK} &Fine-tune &86.93 &20.81 &76.06 &117.47 &26.42 &69.61 &50.19 &25.11 &71.11 &183.30 &72.26 \\
&PGD &86.24 &\underline{37.18} &\underline{56.89} &140.17 &33.33 &61.36 &53.01 &\textbf{40.28} &\underline{53.30} &\underline{194.52} &\underline{57.18} \\
&FreeLB &\textbf{88.79} &28.05 &68.41 &125.13 &31.80 &64.19 &52.57 &30.98 &65.11 &188.02 &65.90 \\
&InfoBERT &\underline{88.73} &26.97 &69.61 &125.24 &30.68 &65.43 &51.62 &28.76 &67.59 &186.43 &67.54 \\
&TAVAT &87.85 &35.51 &59.58 &\textbf{147.97} &33.80 &61.53 &53.65 &35.55 &59.54 &191.88 &60.21 \\
&DNE &82.13 &29.49 &63.94 &88.77 &19.38 &76.31 &\underline{54.71} &25.53 &68.91 &141.88 &69.80 \\
&SAFER &86.85 &27.78 &68.01 &135.33 &\underline{34.10} &60.74 &52.78 &33.57 &61.35 &193.89 &63.37 \\
&Flooding-X &86.53 &24.75 &71.40 &119.77 &23.99 &72.27 &48.19 &26.68 &69.16 &184.09 &70.95 \\
&ALS &86.28 &29.17 &66.19 &125.74 &27.90 &67.66 &47.90 &28.11 &67.42 &186.23 &67.09 \\
&AdvFooler &85.73 &30.91 &63.94 &140.97 &33.79 &\underline{60.59} &53.46 &34.49 &59.77 &194.49 &61.43 \\
% \cdashlinelr{2-13}
&\textbf{\textsc{PuRe} (Ours)} &84.32 &\textbf{38.67} &\textbf{54.12} &\underline{143.56} &\textbf{38.50} &\textbf{54.32} &\textbf{56.74} &\underline{39.58} &\textbf{53.04} &\textbf{195.34} &\textbf{53.85} \\

\bottomrule
\end{tabular}
}
\caption{\label{tab:main_v3}
Adversarial robustness results with different baselines. \textbf{Bold}: the best. \underline{Underline}: the second best.}
\end{table*}

%% file: tables/top_k_pc.tex
\begin{table}[!t]\centering
\resizebox{\linewidth}{!}{
% \scriptsize
\begin{tabular}{lrrrrr}\toprule
\multirow{2}{*}{\textbf{Setting}} &\multirow{2}{*}{\textbf{\textsc{Acc}}$\uparrow$} &\multicolumn{3}{c}{\textbf{\textsc{Aua}}$\uparrow$} \\\cmidrule{3-5}
& &\textbf{TextFooler} &\textbf{TextBugger} &\textbf{PWWS} \\\midrule
Fine-tune &93.47 &5.66 &34.71 &16.91 \\
\midrule
\textsc{PuRe} & & & & \\
\quad Remove top-1 PC &91.43 &9.06 &34.27 &16.91 \\
\quad Remove top-1 to top-3 PCs &91.16 &13.18 &39.81 &21.86 \\
\quad Remove top-1 to top-5 PCs &87.64 &13.73 &39.32 &23.56 \\
\bottomrule
\end{tabular}
}
\caption{\label{tab:remove_top_k_pc}
Ablation study of removing more than the top-1 principal component for Qwen2.5 on SST2 test set. Removing more than just the top-1 principal component can promote a more isotropic embedding space and improve after-attack accuracy; however, excessive removal may degrade performance on clean examples.
%\red{Removing more than the top-1 component can lead to a more isotropic embedding space, improving after-attack accuracy, though excessive removal diminishes clean accuracy.}
}
\end{table}

%% file: tables/reasoning.tex
\begin{table}[!t]\centering
\resizebox{\linewidth}{!}{
\scriptsize
\begin{tabular}{ccrrrrr}\toprule
\multirow{2}{*}{\textbf{Dataset}} &\multirow{2}{*}{\textbf{Defence}} &\multirow{2}{*}{\textbf{\textsc{Acc}}$\uparrow$} &\multicolumn{3}{c}{\textbf{TextFooler}} \\\cmidrule{4-6}
& & &\textbf{\textsc{Aua}}$\uparrow$ &\textbf{\textsc{Asr}}$\downarrow$ &\textbf{\textsc{AvgQ}}$\uparrow$ \\\midrule
\multirow{4}{*}{SIQA} &Fine-tune &61.51 &3.68 &94.01 &41.77 \\
&Flooding-X &61.07 &3.64 &94.04 &41.61 \\
&ALS &\textbf{61.76} &4.01 &93.51 &42.19 \\
&\textbf{\textsc{PuRe} (Ours)} &59.57 &\textbf{11.62} &\textbf{80.50} &\textbf{51.95} \\
\midrule
\multirow{4}{*}{CSQA} &Fine-tune &57.17 &3.28 &94.26 &25.87 \\
&Flooding-X &57.98 &3.64 &93.72 &24.33 \\
&ALS &\textbf{58.11} &4.79 &91.76 &26.76 \\
&\textbf{\textsc{PuRe} (Ours)} &55.96 &\textbf{7.61} &\textbf{86.40} &\textbf{28.19} \\
\bottomrule
\end{tabular}
}
\caption{\label{tab:reasoning}
The experiment results on the commonsense reasoning tasks using BERT-based model.}
\end{table}

%% file: tables/intra_sim.tex
\begin{table}[!tp]%[!htp]
\centering
% \scriptsize
\scalebox{0.7}{
\begin{tabular}{lccc}\toprule
\textbf{Model} &\textbf{Fine-tune} &\textbf{\textsc{PuRe} (Pre)} & \textbf{\textsc{PuRe} (Post)} \\\midrule
% \multicolumn{3}{l}{\textit{Intra-Sentence Similarity}} \\
Unadjusted & 0.794 & 0.592 & 0.895 \\
- Anisotropy Estimates & 0.129 & 0.002 & 0.008 \\
= Adjusted & 0.665 & 0.590 & 0.887 \\

\bottomrule
\end{tabular}
}
\caption{Intra-sentence similarity score of last hidden layer, with vanilla fine-tuning and \textsc{PuRe}. For \textsc{PuRe}, we measure both pre-\textsc{PuRe} layer and post-\textsc{PuRe} layer.
}
\label{tab:intra-sim}
\end{table}

%% file: tables/pcr_pfsa.tex
\begin{table}[!t]\centering
\resizebox{\linewidth}{!}{
% \scriptsize
\begin{tabular}{cccrrrrr}\toprule
\multirow{2}{*}{\textbf{Dataset}} &\multirow{2}{*}{\textbf{PCR}} &\multirow{2}{*}{\textbf{PFSA}} &\multirow{2}{*}{\textbf{\textsc{Acc}}$\uparrow$} &\multicolumn{3}{c}{\textbf{\textsc{Aua}}$\uparrow$} \\\cmidrule{5-7}
& & & &\textbf{TextFooler} &\textbf{TextBugger} &\textbf{PWWS} \\\midrule
\multirow{3}{*}{SST2} & &\cmark &\textbf{91.98} &16.09 &40.20 &21.75 \\
&\cmark & &91.71 &10.27 &35.15 &16.75 \\
&\cmark &\cmark &90.88 &\textbf{30.37} &\textbf{47.17} &\textbf{36.02} \\
\midrule
\multirow{3}{*}{MRPC} & &\cmark &83.48 &12.35 &12.35 &15.42 \\
&\cmark & &\textbf{84.12} &7.83 &10.32 &11.65 \\
&\cmark &\cmark &82.20 &\textbf{17.22} &\textbf{16.29} &\textbf{18.55} \\
\bottomrule
\end{tabular}
}
\caption{\label{tab:pcr_pfsa}
Ablation study on PCR and PFSA modules.}
\end{table}

% \begin{table}[!t]\centering
% \resizebox{\linewidth}{!}{
% % \scriptsize
% \begin{tabular}{lrrrrrrrr}\toprule
% \multirow{2}{*}{\textbf{Dataset}} &\multirow{2}{*}{\textbf{PCR}} &\multirow{2}{*}{\textbf{PFSA}} &\multicolumn{2}{c}{\textbf{TextFooler}} &\multicolumn{2}{c}{\textbf{TextBugger}} &\multicolumn{2}{c}{\textbf{PWWS}} \\\cmidrule{4-9}
% & & &\textbf{\textsc{Aua}}$\uparrow$ &\textbf{\textsc{Pdr}}$\downarrow$ &\textbf{\textsc{Aua}}$\uparrow$ &\textbf{\textsc{Pdr}}$\downarrow$ &\textbf{\textsc{Aua}}$\uparrow$ &\textbf{\textsc{Pdr}}$\downarrow$ \\\midrule
% \multirow{3}{*}{SST2} & &\cmark &16.09 &82.51 &40.20 &56.29 &21.75 &76.35 \\
% &\cmark & &10.27 &88.80 &35.15 &61.67 &16.75 &81.74 \\
% &\cmark &\cmark &\textbf{29.65} &\textbf{67.69} &\textbf{46.51} &\textbf{49.31} &\textbf{33.94} &\textbf{63.01} \\
% \midrule
% \multirow{3}{*}{MRPC} & &\cmark &12.35 &85.21 &12.35 &85.21 &15.42 &81.53 \\
% &\cmark & &7.83 &90.69 &10.32 &87.73 &11.65 &86.15 \\
% &\cmark &\cmark &\textbf{17.22} &\textbf{79.05} &\textbf{16.29} &\textbf{80.18} &\textbf{18.55} &\textbf{77.43} \\
% \bottomrule
% \end{tabular}
% }
% \caption{\label{tab:pcr_pfsa}
% Ablation study on PCR and PFSA modules on the SST2 and MRPC datasets.}
% \end{table}

%% file: tables/rand_v2.tex
\begin{table}[!t]\centering
\resizebox{\linewidth}{!}{
% \scriptsize
\begin{tabular}{ccrrrrr}\toprule
\multirow{2}{*}{\textbf{Dataset}} &\multirow{2}{*}{\textbf{Randomised}} &\multirow{2}{*}{\textbf{\textsc{Acc}}$\uparrow$} &\multicolumn{3}{c}{\textbf{\textsc{Aua}}$\uparrow$} \\\cmidrule{4-6}
& & &\textbf{TextFooler} &\textbf{TextBugger} &\textbf{PWWS} \\\midrule
\multirow{2}{*}{SST2} &SVD &90.88 &24.49 &44.59 &30.20 \\
&rSVD &\textbf{91.76} &\textbf{30.37} &\textbf{47.17} &\textbf{36.02} \\
\midrule
\multirow{2}{*}{MRPC} &SVD &75.77 &3.86 &6.42 &5.13 \\
&rSVD &\textbf{82.20} &\textbf{17.22} &\textbf{16.29} &\textbf{18.55} \\
\bottomrule
\end{tabular}
}
\caption{\label{tab:rand_v2}
Impact of randomisation in PCR module on the SST2 and MRPC datasets.}
\end{table}

% \begin{table}[!t]\centering
% \resizebox{\linewidth}{!}{
% % \scriptsize
% \begin{tabular}{ccrrrrrrr}\toprule
% \multirow{2}{*}{\textbf{Dataset}} &\multirow{2}{*}{\textbf{Randomised}} &\multicolumn{2}{c}{\textbf{TextFooler}} &\multicolumn{2}{c}{\textbf{TextBugger}} &\multicolumn{2}{c}{\textbf{PWWS}} &\multirow{2}{*}{\textbf{\textsc{Apdr}}} \\\cmidrule{3-8}
% & &\textbf{\textsc{Aua}}$\uparrow$ &\textbf{\textsc{Pdr}}$\downarrow$ &\textbf{\textsc{Aua}}$\uparrow$ &\textbf{\textsc{Pdr}}$\downarrow$ &\textbf{\textsc{Aua}}$\uparrow$ &\textbf{\textsc{Pdr}}$\downarrow$ & \\\midrule
% \multirow{2}{*}{SST2} &SVD &24.49 &73.05 &44.59 &50.94 &30.20 &66.77 &63.59 \\
% &rSVD &\textbf{30.37} &\textbf{66.59} &\textbf{47.17} &\textbf{48.10} &\textbf{36.02} &\textbf{60.36} &\textbf{58.35} \\
% \midrule
% \multirow{2}{*}{MRPC} &SVD &0.64 &99.16 &4.12 &94.56 &2.43 &96.79 &96.84 \\
% &rSVD &\textbf{17.22} &\textbf{79.05} &\textbf{16.29} &\textbf{80.18} &\textbf{18.55} &\textbf{77.43} &\textbf{78.89} \\
% \bottomrule
% \end{tabular}
% }
% \caption{\label{tab:rand_v2}
% Impact of randomisation in PCR module on the SST2 and MRPC datasets.}
% \end{table}

%% file: tables/runtime.tex
\begin{table}[!t]
\centering
\resizebox{\linewidth}{!}{
% \begin{tabular}{lcc}\toprule
% \textbf{Method} &\textbf{Train} &\textbf{Inference} \\\midrule
% Fine-tune &$1.0$ &$1.0$ \\
% PGD \cite{madry2019deep} &$\times 3.3$ &$\times 1.0$ \\
% FreeLB \cite{Zhu2020FreeLB:} &$\times 2.6$ &$\times 1.0$ \\
% InfoBERT \cite{wang2020infobert} &$\times 3.8$ &$\times 1.0$ \\
% TAVAT \cite{li2020tavat} &$\times 1.6$ &$\times 1.0$ \\
% SAFER \cite{ye2020safer} &$\times 1.1$ &$\times 1.0$ \\
% DNE \cite{zhou-etal-2021-defense} &$\times 2.5$ &$\times 3.0$ \\
% Flooding-X \cite{liu2022flooding} &$\times 1.0$ &$\times 1.0$ \\
% ALS \cite{yang-etal-2023-domain} &$\times 1.1$ &$\times 1.0$ \\
% AdvFooler \cite{hoang2024fooling} &$\times 1.0$ &$\times 1.5$ \\
% \textsc{PuRe} (w/ SVD) &$\times 1.9$ &$\times 1.5$ \\
% \textsc{PuRe} (w/ rSVD) &$\times 1.2$ &$\times 1.1$ \\
% \bottomrule
% \end{tabular}
\begin{tabular}{lccc}\toprule
\textbf{Method} &\textbf{Train}$\downarrow$ &\textbf{Inference}$\downarrow$ &\textbf{\(\Delta\)\textsc{Apdr}$\uparrow$} \\\midrule
Fine-tune &$1.0$ &$1.0$ &- \\
PGD \cite{madry2019deep} &$\times 3.3$ &$\times 1.0$ &$10.82$ \\
FreeLB \cite{Zhu2020FreeLB:} &$\times 2.6$ &$\times 1.0$ &$12.07$ \\
InfoBERT \cite{wang2020infobert} &$\times 3.8$ &$\times 1.0$ &$4.54$ \\
TAVAT \cite{li2020tavat} &$\times 1.6$ &$\times 1.0$ &$10.43$ \\
SAFER \cite{ye2020safer} &$\times 1.1$ &$\times 1.0$ &$16.99$ \\
DNE \cite{zhou-etal-2021-defense} &$\times 2.5$ &$\times 3.0$ &$0.45$ \\
Flooding-X \cite{liu2022flooding} &$\times 1.0$ &$\times 1.0$ &$3.81$ \\
ALS \cite{yang-etal-2023-domain} &$\times 1.1$ &$\times 1.0$ &$3.70$ \\
AdvFooler \cite{hoang2024fooling} &$\times 1.0$ &$\times 1.5$ &$3.71$ \\
\textsc{PuRe} (w/ SVD) &$\times 1.9$ &$\times 1.5$ &$13.66$ \\
\textsc{PuRe} (w/ rSVD) &$\times 1.2$ &$\times 1.1$ &$17.17$ \\
\bottomrule
\end{tabular}
}
\caption{Runtime comparison of \textsc{PuRe} and baseline methods on the MRPC dataset, with $\Delta$\textsc{Apdr} indicating the absolute drop in \textsc{Apdr} relative to the fine-tuning baseline.}
\label{tab:runtime}
%\label{tab:runtime}Run time comparison of \textsc{PuRe} and baselines based on the MRPC dataset, \red{along with $\Delta$\textsc{Apdr} representing the absolute drop in \textsc{Apdr} compared to the fine-tune baseline.}
\end{table}

%% file: tables/stats_full.tex
\begin{table*}[!htp]\centering
% \resizebox{\linewidth}{!}{
\scriptsize
\begin{tabular}{lrrrrrrr}\toprule
\textbf{Dataset} &\textbf{Task} &\textbf{Classes} &\textbf{Train} &\textbf{Validation} &\textbf{Test} &\textbf{Label Distribution$^\ast$} \\\midrule
SST2 &Sentiment Analysis &2 &6920 &872 &1821 &Approx. Equal \\
SUBJ &Subjectivity Status &2 &8000 &- &2000 &Approx. Equal \\
CR &Sentiment Analysis &2 &3394 &- &376 &Approx. Equal \\
MR &Sentiment Analysis &2 &8530 &1066 &1066 &Equal \\
MRPC &Paraphrase Identification &2 &3668 &408 &1725 &Approx. Equal \\
SICK &Textual Entailment &3 &4439 &495 &4906 &Approx. Equal \\
SIQA &Commonsense Reasoning &3 &33410 &1954 &- &Approx. Equal \\
CSQA &Commonsense Reasoning &5 &9741 &1221 &- &Approx. Equal \\
\bottomrule
\end{tabular}
% }
\caption{\label{tab:statistics}Statistics of datasets. $^\ast$ Distribution of the examples across classes in validation/test set.}
\end{table*}

%% file: tables/model_card.tex
\begin{table}[!t]\centering
\resizebox{\linewidth}{!}{
% \scriptsize
\begin{tabular}{ccr}\toprule
\textbf{Model} &\textbf{Checkpoint} &\textbf{Params} \\\midrule
BERT-base &google-bert/bert-base-uncased &109M \\
BERT-large &google-bert/bert-large-uncased &335M \\
RoBERTa-base &FacebookAI/roberta-base &124M \\
RoBERTa-large &FacebookAI/roberta-large &355M \\
DeBERTa-base &microsoft/deberta-v3-base &184M \\
DeBERTa-large &microsoft/deberta-v3-large &435M \\
BGE-base &BAAI/bge-base-en-v1.5 &109M \\
BGE-large &BAAI/bge-large-en-v1.5 &335M \\
GIST-base &avsolatorio/GIST-Embedding-v0 &109M \\
GIST-large &avsolatorio/GIST-large-Embedding-v0 &335M \\
OPT-base &facebook/opt-125m &125M \\
OPT-large &facebook/opt-350m &350M \\
Qwen2.5-0.5B &Qwen/Qwen2.5-0.5B &494M \\
\bottomrule
\end{tabular}
}
\caption{\label{tab:model_card}
PLMs checkpoints from HuggingFace Hub. \textbf{Model}: Lists the names of different PLMs. \textbf{Checkpoint}: Specifies the HuggingFace checkpoint name with each model. \textbf{Params}: Indicates the number of parameters in each model (in millions, denoted by ``M'').}
\end{table}

%% file: tables/architecture_full.tex
\begin{table*}[!t]\centering
\resizebox{\linewidth}{!}{
\scriptsize
\begin{tabular}{cccrrrrrrrrrrrr}\toprule
\multirow{2}{*}{\textbf{Model}} &\multirow{2}{*}{\textbf{Params}} &\multirow{2}{*}{\textbf{Defence}} &\multirow{2}{*}{\textbf{\textsc{Acc}}$\uparrow$} &\multicolumn{3}{c}{\textbf{TextFooler}} &\multicolumn{3}{c}{\textbf{TextBugger}} &\multicolumn{3}{c}{\textbf{PWWS}} &\multirow{2}{*}{\textbf{\textsc{Apdr}}$\downarrow$} \\\cmidrule{5-13}
& & & &\textbf{\textsc{Aua}}$\uparrow$ &\textbf{\textsc{Asr}}$\downarrow$ &\textbf{\textsc{AvgQ}}$\uparrow$ &\textbf{\textsc{Aua}}$\uparrow$ &\textbf{\textsc{Asr}}$\downarrow$ &\textbf{\textsc{AvgQ}}$\uparrow$ &\textbf{\textsc{Aua}}$\uparrow$ &\textbf{\textsc{Asr}}$\downarrow$ &\textbf{\textsc{AvgQ}}$\uparrow$ & \\\midrule

\multirow{2}{*}{BERT-base} &\multirow{2}{*}{109M} &Fine-tune &92.09 &6.32 &93.14 &87.07 &30.37 &67.02 &41.19 &13.78 &85.03 &127.57 &81.73 \\
& &\textbf{\textsc{PuRe} (Ours)} &90.88 &30.37 &66.59 &134.01 &47.17 &48.10 &58.52 &36.02 &60.36 &139.97 &58.35 \\
\multirow{2}{*}{BERT-large} &\multirow{2}{*}{335M} &Fine-tune &93.08 &8.84 &90.50 &98.13 &31.63 &66.02 &43.09 &16.69 &82.06 &130.93 &79.53 \\
& &\textbf{\textsc{PuRe} (Ours)} &91.71 &29.87 &67.43 &143.41 &49.75 &45.75 &56.77 &39.21 &57.25 &144.44 &56.81 \\

\midrule
\multirow{2}{*}{RoBERTa-base} &\multirow{2}{*}{124M} &Fine-tune &94.84 &5.88 &93.80 &93.08 &33.06 &65.14 &43.22 &14.61 &84.60 &132.84 &81.18 \\
& &\textbf{\textsc{PuRe} (Ours)} &92.97 &16.14 &82.63 &110.63 &42.78 &53.99 &49.29 &23.45 &74.78 &135.82 &70.47 \\
\multirow{2}{*}{RoBERTa-large} &\multirow{2}{*}{355M} &Fine-tune &95.17 &10.43 &89.04 &109.37 &41.85 &56.03 &46.48 &20.92 &78.02 &138.01 &74.36 \\
& &\textbf{\textsc{PuRe} (Ours)} &94.67 &16.80 &82.25 &115.61 &45.19 &52.26 &52.49 &26.85 &71.64 &138.39 &68.72 \\

\midrule
\multirow{2}{*}{DeBERTa-base} &\multirow{2}{*}{184M} &Fine-tune &95.50 &12.96 &86.43 &108.33 &42.06 &55.95 &44.94 &23.50 &75.39 &137.39 &72.59 \\
& &\textbf{\textsc{PuRe} (Ours)} &94.18 &21.31 &77.38 &122.50 &50.36 &46.53 &51.73 &31.08 &67.00 &139.87 &63.64 \\
\multirow{2}{*}{DeBERTa-large} &\multirow{2}{*}{435M} &Fine-tune &96.54 &9.88 &89.76 &104.12 &48.22 &50.06 &48.57 &22.19 &77.02 &139.09 &72.28 \\
& &\textbf{\textsc{PuRe} (Ours)} &94.95 &37.78 &60.21 &156.56 &67.60 &28.80 &60.92 &46.90 &50.61 &151.65 &46.54 \\

\midrule
\multirow{2}{*}{BGE-base} &\multirow{2}{*}{109M} &Fine-tune &93.03 &7.69 &91.74 &89.63 &29.71 &68.06 &40.84 &14.11 &84.83 &129.43 &81.54 \\
& &\textbf{\textsc{PuRe} (Ours)} &89.35 &19.11 &78.61 &119.54 &46.07 &48.43 &47.84 &28.78 &67.79 &135.83 &64.94 \\
\multirow{2}{*}{BGE-large} &\multirow{2}{*}{335M} &Fine-tune &94.07 &10.27 &89.08 &98.95 &35.86 &61.88 &42.55 &17.41 &81.49 &133.11 &77.48 \\
& &\textbf{\textsc{PuRe} (Ours)} &91.93 &30.81 &66.49 &136.80 &50.03 &45.58 &53.20 &37.51 &59.20 &143.60 &57.09 \\

\midrule
\multirow{2}{*}{GIST-base} &\multirow{2}{*}{109M} &Fine-tune &92.86 &5.82 &93.73 &87.91 &30.64 &67.00 &40.41 &12.58 &86.46 &129.25 &82.40 \\
& &\textbf{\textsc{PuRe} (Ours)} &89.02 &24.22 &72.79 &127.24 &44.26 &50.28 &48.02 &27.40 &69.22 &134.71 &64.10 \\
\multirow{2}{*}{GIST-large} &\multirow{2}{*}{335M} &Fine-tune &94.45 &8.40 &91.10 &96.28 &33.77 &64.24 &42.73 &17.79 &81.16 &133.15 &78.83 \\
& &\textbf{\textsc{PuRe} (Ours)} &92.09 &25.48 &72.33 &129.42 &48.11 &47.76 &52.64 &33.28 &63.86 &139.80 &61.32 \\

\midrule
\multirow{2}{*}{OPT-base} &\multirow{2}{*}{125M} &Fine-tune &92.48 &4.56 &95.07 &83.21 &28.50 &69.18 &40.71 &12.63 &86.34 &127.52 &83.53 \\
& &\textbf{\textsc{PuRe} (Ours)} &86.44 &13.67 &84.18 &104.09 &35.42 &59.02 &47.84 &21.09 &75.60 &129.93 &72.93 \\
\multirow{2}{*}{OPT-large} &\multirow{2}{*}{331M} &Fine-tune &93.25 &4.72 &94.94 &83.07 &30.42 &67.37 &40.74 &11.97 &87.16 &128.90 &83.16 \\
& &\textbf{\textsc{PuRe} (Ours)} &90.28 &16.31 &81.93 &112.57 &39.87 &55.84 &48.68 &24.99 &72.32 &135.37 &70.03 \\

\midrule
\multirow{2}{*}{Qwen2.5-0.5B} &\multirow{2}{*}{494M} &Fine-tune &93.47 &5.66 &93.95 &86.62 &34.71 &62.87 &42.39 &16.91 &81.90 &132.36 &79.57 \\
& &\textbf{\textsc{PuRe} (Ours)} &91.43 &9.06 &90.09 &95.18 &34.27 &62.52 &46.39 &16.91 &81.50 &132.21 &78.04 \\
\bottomrule
\end{tabular}
}
\caption{\label{tab:architectures_full}
Comparison of different model architectures on the SST2 test set.}
\end{table*}

%% file: tables/transferability_full.tex
\begin{table*}[!t]\centering
\resizebox{\linewidth}{!}{
% \scriptsize
\begin{tabular}{cccrrrrrrrrrrrr}\toprule
\multicolumn{2}{c}{\textbf{Dataset}} &\multirow{2}{*}{\textbf{Method}} &\multirow{2}{*}{\textbf{\textsc{Acc}}$\uparrow$} &\multicolumn{3}{c}{\textbf{TextFooler}} &\multicolumn{3}{c}{\textbf{TextBugger}} &\multicolumn{3}{c}{\textbf{PWWS}} &\multirow{2}{*}{\textbf{\textsc{Apdr}}$\downarrow$} \\\cmidrule{1-2}\cmidrule{5-13}
\textbf{Source} &\textbf{Target} & & &\textbf{\textsc{Aua}}$\uparrow$ &\textbf{\textsc{Asr}}$\downarrow$ &\textbf{\textsc{AvgQ}}$\uparrow$ &\textbf{\textsc{Aua}}$\uparrow$ &\textbf{\textsc{Asr}}$\downarrow$ &\textbf{\textsc{AvgQ}}$\uparrow$ &\textbf{\textsc{Aua}}$\uparrow$ &\textbf{\textsc{Asr}}$\downarrow$ &\textbf{\textsc{AvgQ}}$\uparrow$ & \\\midrule
\multirow{11}{*}{CR} &\multirow{11}{*}{SST2} &Fine-tune &84.24 &3.95 &95.31 &77.96 &23.39 &72.23 &38.36 &8.90 &89.44 &121.47 &85.66 \\
& &PGD &82.76 &6.70 &91.90 &89.90 &25.81 &68.81 &41.17 &14.22 &82.81 &\underline{129.04} &81.17 \\
& &FreeLB &83.09 &4.89 &94.12 &86.03 &26.58 &68.01 &40.29 &11.70 &85.92 &126.51 &82.68 \\
& &InfoBERT &84.73 &5.33 &93.71 &83.79 &24.55 &71.03 &39.34 &11.81 &86.07 &125.50 &83.60 \\
& &TAVAT &81.71 &7.25 &91.13 &93.81 &26.80 &67.20 &41.16 &15.05 &81.59 &129.37 &79.97 \\
& &DNE &\underline{85.11} &\underline{14.39} &\underline{83.09} &\underline{111.46} &27.33 &67.86 &\textbf{51.69} &\textbf{28.00} &\underline{67.04} &104.84 &\underline{72.66} \\
& &SAFER &82.67 &8.22 &90.05 &104.52 &\underline{31.83} &\underline{61.60} &50.02 &23.83 &71.25 &104.93 &74.30 \\
& &Flooding-X &84.79 &2.47 &97.09 &74.31 &21.09 &75.13 &37.02 &8.57 &89.90 &122.23 &87.37 \\
& &ALS &\textbf{85.28} &6.21 &92.72 &86.65 &28.01 &67.16 &40.76 &12.19 &85.71 &123.80 &81.86 \\
& &AdvFooler &79.91 &7.32 &90.84 &90.44 &28.49 &64.35 &45.54 &12.02 &84.96 &127.77 &80.05 \\
& &\textbf{\textsc{PuRe} (Ours)} &77.65 &\textbf{21.53} &\textbf{72.28} &\textbf{123.00} &\textbf{39.87} &\textbf{48.66} &\underline{50.12} &\underline{27.79} &\textbf{64.21} &\textbf{135.51} &\textbf{61.72} \\

\midrule
\multirow{11}{*}{SST2} &\multirow{11}{*}{CR} &Fine-tune &\textbf{85.37} &2.13 &97.51 &76.36 &28.46 &66.67 &34.31 &9.04 &89.41 &129.23 &84.53 \\
& &PGD &80.85 &8.78 &89.14 &87.87 &28.99 &64.14 &35.25 &18.09 &77.63 &134.22 &76.97 \\
& &FreeLB &84.04 &6.65 &92.09 &85.03 &27.93 &66.77 &35.61 &14.63 &82.59 &133.03 &80.48 \\
& &InfoBERT &83.78 &5.05 &93.97 &83.83 &25.27 &69.84 &35.05 &14.36 &82.86 &132.41 &82.22 \\
& &TAVAT &82.18 &8.78 &89.32 &90.23 &26.60 &67.64 &36.67 &18.09 &77.99 &\underline{135.77} &78.32 \\
& &DNE &72.53 &\underline{13.46} &\textbf{81.58} &103.14 &25.82 &64.39 &\textbf{48.96} &19.23 &73.48 &110.01 &73.15 \\
& &SAFER &\underline{85.16} &10.16 &88.06 &\textbf{105.68} &\textbf{35.44} &\textbf{58.12} &\underline{47.16} &\textbf{23.08} &\textbf{72.73} &109.94 &\underline{72.97} \\
& &Flooding-X &82.45 &4.52 &94.52 &80.03 &27.66 &66.45 &35.39 &13.03 &84.19 &132.11 &81.72 \\
& &ALS &83.78 &4.52 &94.60 &82.50 &29.79 &64.44 &35.03 &9.04 &89.21 &127.33 &82.75 \\
& &AdvFooler &80.02 &9.93 &87.59 &99.48 &27.71 &65.37 &45.59 &17.39 &78.27 &130.21 &77.08 \\
& &\textbf{\textsc{PuRe} (Ours)} &82.18 &\textbf{14.89} &\underline{81.88} &\underline{104.05} &\underline{33.78} &\underline{58.90} &41.77 &\underline{22.07} &\underline{73.14} &\textbf{136.67} &\textbf{71.31} \\

\midrule
\multirow{11}{*}{MR} &\multirow{11}{*}{SST2} &Fine-tune &93.68 &10.60 &88.69 &103.33 &35.69 &61.90 &44.93 &18.12 &80.66 &130.63 &77.08 \\
& &PGD &94.34 &17.02 &81.96 &\textbf{130.29} &48.11 &49.01 &49.54 &\underline{33.94} &\underline{64.03} &\textbf{144.01} &65.00 \\
& &FreeLB &94.73 &11.53 &87.83 &114.08 &42.01 &55.65 &47.41 &23.39 &75.30 &138.34 &72.93 \\
& &InfoBERT &94.73 &11.09 &88.29 &115.44 &40.91 &56.81 &47.05 &25.92 &72.64 &139.05 &72.58 \\
& &TAVAT &94.34 &18.07 &80.85 &125.73 &45.96 &51.28 &48.71 &33.66 &64.32 &\underline{143.25} &65.48 \\
& &DNE &93.44 &10.11 &89.18 &92.50 &23.06 &75.27 &45.99 &21.50 &76.99 &105.23 &80.48 \\
& &SAFER &95.22 &\underline{21.28} &\underline{77.65} &141.70 &\textbf{50.58} &\textbf{46.91} &\textbf{58.17} &\textbf{44.11} &\textbf{53.68} &107.11 &\textbf{59.41} \\
& &Flooding-X &94.01 &5.05 &94.63 &87.04 &28.67 &69.51 &41.86 &13.12 &86.04 &128.63 &83.39 \\
& &ALS &\underline{95.00} &8.79 &90.75 &105.76 &39.98 &57.92 &44.38 &17.08 &82.02 &132.29 &76.90 \\
& &AdvFooler &93.56 &15.78 &83.13 &120.01 &37.82 &59.58 &51.34 &27.90 &70.18 &138.33 &70.96 \\
& &\textbf{\textsc{PuRe} (Ours)} &\textbf{95.72} &\textbf{25.15} &\textbf{73.72} &\underline{130.26} &\underline{49.97} &\underline{47.79} &\underline{57.10} &31.85 &66.72 &138.69 &\underline{62.74} \\

\midrule
\multirow{11}{*}{SST2} &\multirow{11}{*}{MR} &Fine-tune &88.84 &6.00 &93.24 &94.82 &27.39 &69.17 &47.35 &13.98 &84.27 &138.47 &82.23 \\
& &PGD &89.02 &12.20 &86.30 &123.84 &36.68 &58.80 &52.28 &25.98 &70.81 &\underline{149.28} &71.97 \\
& &FreeLB &88.74 &10.51 &88.16 &119.57 &36.12 &59.30 &53.50 &22.98 &74.10 &148.97 &73.85 \\
& &InfoBERT &88.84 &7.69 &91.34 &111.08 &33.02 &62.83 &49.83 &19.79 &77.72 &145.65 &77.30 \\
& &TAVAT &88.84 &12.66 &85.74 &\underline{132.17} &\underline{39.49} &\underline{55.54} &53.71 &26.92 &69.69 &\textbf{151.60} &70.32 \\
& &DNE &82.87 &\underline{14.27} &\underline{82.78} &124.83 &27.33 &66.82 &\underline{57.06} &\underline{28.54} &\underline{65.44} &114.46 &71.68 \\
& &SAFER &\textbf{89.82} &13.32 &85.17 &\textbf{136.63} &\textbf{44.48} &\textbf{50.37} &\textbf{61.84} &\textbf{38.06} &\textbf{57.49} &115.95 &\textbf{64.34} \\
& &Flooding-X &88.56 &3.00 &96.61 &87.24 &24.58 &72.25 &43.43 &12.10 &86.33 &138.73 &85.06 \\
& &ALS &\underline{89.68} &7.13 &92.05 &103.73 &31.80 &64.54 &47.36 &15.85 &82.32 &140.79 &79.64 \\
& &AdvFooler &88.44 &12.99 &85.31 &110.30 &35.29 &60.10 &53.39 &24.10 &72.75 &120.68 &72.72 \\
& &\textbf{\textsc{PuRe} (Ours)} &88.65 &\textbf{18.76} &\textbf{78.84} &119.91 &37.52 &57.67 &55.47 &24.30 &72.59 &146.13 &\underline{69.70} \\
\bottomrule
\end{tabular}
}
\caption{\label{tab:transferability_full}
Experimental results when the models are trained on the \textbf{source} dataset and then transferred to the \textbf{target} dataset for testing.}
\end{table*}

%% file: tables/ablation.tex
\begin{table*}[!t]\centering
\scriptsize
\begin{tabular}{ccrrrrrrrrrrr}\toprule
\multicolumn{2}{c}{\textbf{Hyper-parameter}} &\multirow{2}{*}{\textbf{\textsc{Acc}}$\uparrow$} &\multicolumn{3}{c}{\textbf{TextFooler}} &\multicolumn{3}{c}{\textbf{TextBugger}} &\multicolumn{3}{c}{\textbf{PWWS}} \\\cmidrule{1-2}\cmidrule{4-12}
\textbf{Key} &\textbf{Value} & &\textbf{\textsc{Aua}}$\uparrow$ &\textbf{\textsc{Asr}}$\downarrow$ &\textbf{\textsc{AvgQ}}$\uparrow$ &\textbf{\textsc{Aua}}$\uparrow$ &\textbf{\textsc{Asr}}$\downarrow$ &\textbf{\textsc{AvgQ}}$\uparrow$ &\textbf{\textsc{Aua}}$\uparrow$ &\textbf{\textsc{Asr}}$\downarrow$ &\textbf{\textsc{AvgQ}}$\uparrow$ \\\midrule
$r$ &8 &90.88 &\textbf{30.37} &\textbf{66.59} &\textbf{134.01} &47.17 &48.10 &\textbf{58.52} &\textbf{36.02} &\textbf{60.36} &\textbf{139.97} \\
$r$ &16 &92.15 &27.84 &69.79 &132.42 &\textbf{48.54} &\textbf{47.32} &55.74 &31.96 &65.32 &137.25 \\
$r$ &32 &91.48 &25.70 &71.89 &127.82 &45.80 &49.91 &57.49 &31.91 &65.11 &137.30 \\
\midrule
$\alpha$ &0.5 &91.43 &17.46 &80.89 &105.92 &39.21 &57.09 &50.57 &23.94 &73.80 &132.40 \\
$\alpha$ &1.0 &91.32 &23.28 &74.50 &121.36 &43.82 &52.01 &53.86 &28.56 &68.73 &135.16 \\
$\alpha$ &1.5 &90.88 &\textbf{30.37} &\textbf{66.59} &\textbf{134.01} &\textbf{47.17} &\textbf{48.10} &\textbf{58.52} &\textbf{36.02} &\textbf{60.36} &\textbf{139.97} \\
$\alpha$ &2.0 &90.50 &20.48 &77.37 &113.14 &38.77 &57.16 &49.73 &26.74 &70.45 &133.81 \\
$\alpha$ &2.5 &90.39 &18.67 &79.34 &114.26 &41.85 &53.71 &51.17 &26.25 &70.96 &134.17 \\
\bottomrule
\end{tabular}
\caption{Impact of $r$ (a sampling parameter indicating the number of Gaussian random vectors mentioned in \S\ref{sec:rsvd}) and $\alpha$ (a scaling factor used to enhance feature expression mentioned in \citet{zhai2023simple}) with BERT-based model on the SST2 dataset.}\label{tab:ablation}
\end{table*}